\documentclass[10pt,conference]{IEEEtran}

\usepackage{tikz}
\usepackage{amsmath}
\usepackage{mathtools,bm}

\usepackage{booktabs}
\usepackage{filecontents}
\usepackage{bbding}

\usepackage{graphicx}
\usepackage[many]{tcolorbox}

\usepackage{mathtools,amssymb,latexsym,amsfonts,stmaryrd}
\usepackage{pbox}
\usepackage{xfrac}
\usepackage{booktabs}
\usepackage{xcolor,colortbl}
\usepackage{threeparttable}
\usepackage{amsthm}
\usepackage{scalerel}

\usepackage{multirow}
\usepackage{tikz}
\usepackage{mathrsfs}
\usepackage{mathpartir}
\usepackage[ruled,linesnumbered,vlined,noend]{algorithm2e}
\usepackage{caption}
\usepackage{url}
\usepackage{syntax}
\usepackage{framed}
\usepackage[noend]{algpseudocode}
\usepackage{flushend}
\usepackage{listings}
\usepackage{moresize}
\usepackage{wrapfig}
 
\usepackage{seqsplit}
\usepackage{alltt}
\usepackage{xspace}
\usepackage{enumitem}
\usepackage{pifont}

\DeclareMathAlphabet{\mathcal}{OMS}{cmsy}{m}{n}

\usepackage{amsmath, listings, amsthm, amssymb, proof, xspace}

\usepackage{array}
\newcommand{\PreserveBackslash}[1]{\let\temp=\\#1\let\\=\temp}
\newcolumntype{C}[1]{>{\PreserveBackslash\centering}p{#1}}
\newcolumntype{R}[1]{>{\PreserveBackslash\raggedleft}p{#1}}
\newcolumntype{L}[1]{>{\PreserveBackslash\raggedright}p{#1}}

\usepackage{thmtools}

\declaretheoremstyle[spaceabove=\topsep,notefont=\normalfont\itshape]{mystyle}

\usepackage[colorlinks]{hyperref}   
\AtBeginDocument{%
  \hypersetup{pdfborder={0 0 1}, urlcolor=.}  
}

\definecolor[named]{ACMBlue}{cmyk}{1,0.1,0,0.1}
\definecolor[named]{ACMYellow}{cmyk}{0,0.16,1,0}
\definecolor[named]{ACMOrange}{cmyk}{0,0.42,1,0.01}
\definecolor[named]{ACMRed}{cmyk}{0,0.90,0.86,0}
\definecolor[named]{ACMLightBlue}{cmyk}{0.49,0.01,0,0}
\definecolor[named]{ACMGreen}{cmyk}{0.20,0,1,0.19}
\definecolor[named]{ACMPurple}{cmyk}{0.55,1,0,0.15}
\definecolor[named]{ACMDarkBlue}{cmyk}{1,0.58,0,0.21}
\hypersetup{
  colorlinks,
  linkcolor=ACMPurple,
  citecolor=ACMPurple,
  urlcolor=ACMDarkBlue,
  filecolor=ACMDarkBlue
}

\usepackage{inconsolata} 
\usepackage{mathptmx} 
\usepackage{biolinum} 

\usepackage{microtype}

\usepackage[noadjust]{cite}

\newcommand{\revise}[2]{{\color{red}{\ifx&#1&\else- #1\fi}} {\color{ForestGreen}{\ifx&#2&\else+ #2\fi}}}%
\renewcommand{\revise}[2]{#2}%

\usetikzlibrary{arrows,patterns, decorations.pathreplacing}

\newcommand{\F}{Fig.}
\newcommand{\E}{Eq.}
\newcommand{\T}{Table}
\renewcommand{\S}{Sec.}
\newcommand{\A}{Alg.}

\newcommand{\ignore}[1]{}

\lstdefinestyle{base}{
  moredelim=**[is][\color{red}]{@}{@},
  escapeinside={<@}{@>}
}

\newcommand{\tool}{\textsc{NLC}}
\newcommand{\nac}{\textsc{NLC}}

\usepackage{amssymb}
\usepackage{pifont}

\newcommand\DejaVuttfamily{%
  \fontfamily{DejaVuSansMono-TLF}\selectfont }

\lstdefinestyle{base}{
  moredelim=**[is][\color{red}]{@}{@},
  escapeinside={<@}{@>}
}

\lstdefinelanguage
   [x64]{Assembler}     
   [x86masm]{Assembler} 
   {morekeywords={CDQE,CQO,CMPSQ,CMPXCHG16B,JRCXZ,LODSQ,MOVSXD, %
                  POPFQ,PUSHFQ,SCASQ,STOSQ,IRETQ,RDTSCP,SWAPGS, %
                  rax,rdx,rcx,rbx,rsi,rdi,rsp,rbp, %
                  r8,r8d,r8w,r8b,r9,r9d,r9w,r9b}} 

\usepackage[compact]{titlesec}

\usepackage{listings}
\usepackage{fancyvrb}
\usepackage{color}
\definecolor{lightgray}{rgb}{.9,.9,.9}
\definecolor{darkgray}{rgb}{.4,.4,.4}
\definecolor{purple}{rgb}{0.65, 0.12, 0.82}
\definecolor{commentgreen}{RGB}{63,127,95}

\colorlet{myPurple}{blue!40!red}
\definecolor{myOrange}{RGB}{255,192,0}

\newcommand{\enc}[1]{$\phi^{*}_{\theta}$}
\newcommand{\dec}[1]{$\psi^{*}_{\theta}$}

\lstdefinelanguage{Solidity}{
  keywords={len,delete,int,void,payable, public, event, contract, typeof, new, true, false, catch, function, return, null, catch, switch, var, if, in, while, do, else, case, break,struct,const,socklen_t,sa_familty_t,char,sockaddr},
  keywordstyle=\color{violet}\bfseries,
  ndkeywords={class, export, boolean, throw, implements, import, this},
  ndkeywordstyle=\color{darkgray}\bfseries,
  identifierstyle=\color{black},
  sensitive=false,
  comment=[l]{//},
  escapeinside={(*@}{@*)},          
  morecomment=[s]{/*}{*/},
  commentstyle=\color{commentgreen}\ttfamily,
  stringstyle=\color{red}\ttfamily,
  morestring=[b]',
  morestring=[b]"
}

\newcommand{\rnum}[1]{\uppercase\expandafter{\romannumeral #1\relax}}

\usepackage{xcolor}

\algnewcommand{\LeftComment}[1]{\Statex \(\triangleright\) #1}

\definecolor{pptbrown}{RGB}{132,60,12}
\definecolor{pptgreen}{RGB}{56,87,35}

\let\OLDthebibliography\thebibliography
\renewcommand\thebibliography[1]{
  \OLDthebibliography{#1}
  \setlength{\parskip}{0pt}
  \setlength{\itemsep}{0pt plus 0.1ex}
}

\definecolor{pptgreen}{RGB}{84,130,53}
\definecolor{pptred}{RGB}{176,35,24}

\definecolor{pptgreen1}{RGB}{78,173,91}
\definecolor{pptred1}{RGB}{192,0,0}

\definecolor{pptyellow1}{RGB}{203,195,167}
\definecolor{pptgreen2}{RGB}{184,192,176}

\definecolor{pptred3}{RGB}{192,0,0}
\definecolor{pptyellow3}{RGB}{255,192,0}
\definecolor{pptgreen3}{RGB}{4,216,178}

\definecolor{pptblue}{RGB}{0,176,240}
\definecolor{pptgrey}{RGB}{175,171,171}

\newcommand{\CBrush}{\textcolor[RGB]{84,130,53}{\Checkmark}}
\newcommand{\XBrush}{\textcolor[RGB]{176,35,24}{\XSolidBrush}}
\newcommand{\TriUp}{\textcolor[RGB]{0,112,192}{$\Diamond$}}

\newlength{\dpcircle}
\newlength{\rcircle}
\newlength{\dcircle}
\newcommand{\docircle}[4]{%
  \setlength{\dpcircle}{\dp\strutbox}%
  \setlength{\dcircle}{\dpcircle}%
  \addtolength{\dcircle}{\ht\strutbox}%
  \setlength{\rcircle}{0.5\dcircle}%
  \setlength{\unitlength}{1sp}%
  \begin{picture}(\number\dcircle,0)
    \color{#1}
    \put(\number\rcircle,\number\dpcircle){\circle*{\number\dcircle}}
    \color{#2}
    \put(\number\rcircle,\number\dpcircle){\circle{\number\dcircle}}
    \put(\number\rcircle,0){\makebox[0pt]{\textcolor{#3}{#4}}}
  \end{picture}%
}

\newcommand{\Grid}[1]{
  \begin{tikzpicture}
  \node[rectangle,
      draw = black,
      text = olive,
      fill = #1,
      minimum width = 2pt, 
      minimum height = 2pt] (r) at (0,0) {};   
  \end{tikzpicture}
}

\newcommand{\CircleOne}{{\footnotesize\docircle{black}{white}{white}{1}}}
\newcommand{\CircleTwo}{{\footnotesize\docircle{black}{white}{white}{2}}}
\newcommand{\CircleThree}{{\footnotesize\docircle{black}{white}{white}{3}}}
\newcommand{\CircleFour}{{\footnotesize\docircle{black}{white}{white}{4}}}
\newcommand{\CircleFive}{{\footnotesize\docircle{black}{white}{white}{5}}}
\newcommand{\CircleSix}{{\footnotesize\docircle{black}{white}{white}{6}}}
\newcommand{\CircleSeven}{{\footnotesize\docircle{black}{white}{white}{7}}}
\newcommand{\CircleEight}{{\footnotesize\docircle{black}{white}{white}{8}}}

\IEEEoverridecommandlockouts
\begin{document}

\title{Revisiting Neuron Coverage for DNN Testing: A Layer-Wise and
Distribution-Aware Criterion\IEEEauthorrefmark{1}\thanks{\IEEEauthorrefmark{1}The extended version of the ICSE 2023 paper~\cite{yuan2023revisit}}}

\author{
  \IEEEauthorblockN{Yuanyuan Yuan, Qi Pang, and Shuai Wang\IEEEauthorrefmark{2}\thanks{\IEEEauthorrefmark{2} Corresponding author}}
  \IEEEauthorblockA{The Hong Kong University of Science and Technology,
  Hong Kong, China \\
  \tt \{yyuanaq, qpangaa, shuaiw\}@cse.ust.hk}
}

\maketitle

\thispagestyle{plain}
\pagestyle{plain}

\begin{abstract}

  Various deep neural network (DNN) coverage criteria have been proposed to
  assess DNN test inputs and steer input mutations. The coverage is
  characterized via neurons having certain outputs, or the discrepancy
  between neuron outputs. Nevertheless, recent research indicates that
  neuron coverage criteria show little correlation with test suite quality.
  
  In general, DNNs approximate \textit{distributions}, by incorporating
  hierarchical layers, to make predictions for inputs. Thus, we
  champion to deduce DNN behaviors based on its approximated distributions
  from a \textit{layer} perspective. A test suite should be assessed using its
  induced layer output distributions. Accordingly, to fully examine DNN behaviors,
  input mutation should be directed toward diversifying the
  approximated distributions.

  This paper summarizes eight design requirements for DNN coverage criteria,
  taking into account distribution properties and practical concerns. We then
  propose a new criterion, \textsc{NeuraL Coverage} (\tool), that satisfies all
  design requirements. \tool\ treats a single DNN layer as the basic
  computational unit (rather than a single neuron) and captures
  four critical properties of neuron output distributions. Thus, \tool\ accurately
  describes how DNNs comprehend inputs via approximated distributions.
  We demonstrate that \tool\ is significantly correlated with the diversity of a
  test suite across a number of tasks (classification and generation) and data
  formats (image and text). Its capacity to discover DNN prediction errors is
  promising. Test input mutation guided by \tool\ results in a greater quality and
  diversity of exposed erroneous behaviors.

\end{abstract}

\IEEEpeerreviewmaketitle

\section{Introduction}
\label{sec:introduciton}

Recent work has incorporated software testing methodologies to successfully
evaluate DNN~\cite{tian2018deeptest,zhang2018deeproad,wang2020metamorphic}.
However, fundamental understanding of a DNN's mechanism has remained largely
unexplored due to its intrinsic non-linearity. As a result, DNN evaluation is
highly dependent on the test suite's representativeness. Testing criteria are
thus required to characterize a test suite's quality and toward which the input
mutations can comprehensively explore DNN behaviors.

Similar to how coverage criteria are developed in software testing, recent
research proposes several neuron coverage criteria. In principle, the current
criteria are mostly designed to monitor the outputs
\footnote{Often referred to as ``neuron
 activations'' by previous works.}
 of neurons in a DNN model. They argue that each neuron is an
``individual computing unit'' (as the authors put it~\cite{pei2017deepxplore}),
or ``individually encoding a feature''~\cite{ma2018deepgauge}, comparable to a
statement in traditional software. Further, despite neuron outputs are
continuous, existing criteria often first convert neuron outputs into discrete
states (e.g., activated or not) to reduce the modeling complexity. Thus, by
analyzing the discrete states of neurons, the DNN coverage exercised by an input
is measured. Recent works measure the distance between neuron output traces of a
new input and historical inputs to decide the discrepancy of the new input. The
derived coverage is calculated via clusters~\cite{odena2018tensorfuzz} or
discretized buckets over some distance
scores~\cite{kim2019guiding,kim2020reducing}.

However, the key assumption --- \textit{each neuron is an individual computing
  unit} --- often does not hold true for modern DNNs. Given a DNN is composed of
stacked and non-linear layers, the intermediate outputs (also known as latent
representations) from neurons are continuous and highly entangled. Disentangling
intermediate outputs remains a long-lasting
challenge~\cite{lee2018diverse,ren2021interpreting,liu2018exploring}. Recent
research also points out that the
efficacy of a test suite to uncover faults and existing neuron coverage attained
by the test suite are not always positively correlated~\cite{harel2020neuron}.

Establishing proper DNN coverage criteria requires a principled understanding
about how DNN performs computations, which is fundamentally different with
traditional software. A DNN, as a composition of non-linear
functions (i.e., \textit{layers}), extracts task-oriented features
(e.g., classification vs. localization) and then makes decisions. Features, that
are hierarchically propagated by layer outputs, approximate the posterior
distribution of underlying explanatory factors $H$ for an input $x$, namely
$P(H|x)$~\cite{bengio2013representation,bishop1995neural}.
For example, an image classifier generates a set of posterior distributions
$P(h_{i}|x)$ (where $i$ is the layer index and $h_{i} \in H$) with
each $h_{i}$ depicts class-invariant properties of an image from different
aspects (e.g., shape, texture)~\cite{bengio2013representation}. In addition,
the dependability of a DNN is defined as the degree to which the distribution,
approximated on training data, is accurate when applied to real test data.

This work initializes a principled view on designing DNN coverage
criteria, by assessing test inputs regarding their induced layer
output distributions.
We first concretize our focus into eight design requirements. These
requirements characterize the distribution approximated by a DNN from
various aspects, and also take practical constraints (e.g.,
cost, hyper-parameter free) into account. Then, we propose a new coverage
criterion, namely \tool\ (\underline{N}eura\underline{L} \underline{C}overage),
that can fully meet the design requirements. \tool\ is referred to as ``neural
coverage'' rather than neuron coverage since we focus on the overall activity of
groups of neurons (i.e., each layer) and consider relations of neurons
(different with all existing criteria; see comparison in
\S~\ref{subsec:nc}). Our evaluation includes nine DNNs that analyze various
data formats (e.g., image, text) and perform various tasks (e.g.,
classification, generation). We show that \tool\ is substantially correlated
with the diversity and capacity of a test suite to reveal DNN defects.
Meanwhile, by taking \tool\ as the fuzzing feedback, we show that more and
diverse errors can be found, and the mutations guided by \tool\ generate
more natural inputs.
Our contributions are summarized as:

\begin{itemize}[noitemsep,topsep=0pt]
\item
  This work advocates to design DNN testing criteria via a principled
  understanding about how DNN, with its hierarchical layers,
  approximate distributions. We argue that a proper DNN testing
  criterion should model the continuous neuron outputs and neuron entanglement,
  instead of analyzing each neuron individually.

\item We propose eight design requirements by considering the distributions
  approximated by a DNN and practical implementation
  considerations. We accordingly design \tool, satisfying all these
  requirements.

\item We show that \tool\ can better assess the quality of test suites. \tool\ is
  particularly correlated with the diversity and fault-revealing capacity of a
  test suite. Fuzz testing under the guidance of \tool\ can also detect a large
  number of errors, which are shown as diverse and natural-looking.

\end{itemize}

\noindent \textbf{Artifact:}~We released our artifact and supplementary materials
at \url{https://github.com/Yuanyuan-Yuan/NeuraL-Coverage}~\cite{snapshot}.

\section{Preliminary and Overview}
\label{sec:dnn}

DNNs are often explained from the \textit{distribution
approximation} perspective~\cite{bengio2013representation,bishop1995neural,schmidhuber2015deep}.
Following, this section provides a high-level overview of DNNs to motivate \tool. 

\noindent \textbf{``Neural'' and ``Network''.}~The term \textit{neural} in DNN
may be obscure. In cognitive science, neural signifies non-linear functions
(e.g., Sigmoid) in DNNs that simulate how human neurons process
signals~\cite{rosenblatt1958perceptron,block1962perceptron}. DNNs are powerful
at analyzing high-dimensional media data due to their
non-linearity~\cite{bellman1966dynamic,bengio2013representation}.
Neurons in a layer are therefore highly \textit{entangled}, impeding to comprehend
the DNN's internal mechanics. Moreover, outputs of neurons are \textit{continuous}.
The term \textit{network} in DNN refers to topological connections of non-linear
layers. Each layer has a group of neurons cooperating to process certain input
features. The DNN \textit{hierarchy} is established by gradually abstracting
features from shallow to deep layers. For instance, image classifiers commonly
collect shapes in an image at shallow layers and texture at deeper
layers~\cite{geirhos2018imagenettrained,zeiler2014visualizing}. Each
layer (\textit{not} each neuron) serves a minimal computing unit, whose output
distribution encodes its focused features.

Given a collection of data $X$ labeled as $Y$, most popular DNNs can be
categorized as discriminative and generative models, depending on how $X$ and
$Y$ are involved in the \textit{distribution approximation}.

\noindent \textbf{Discriminative Models.}~Deep discriminative models approximate
the conditional probability distribution $P(Y|X)$. A classification model
requires $Y$ as discrete labels, and a regression model (e.g., forecasting)
requires $Y$ as a range of continuous values. For instance, a convolutional
neural network (CNN) can classify images by extracting features (conveyed via
outputs of intermediate neurons) and making decisions in one forward
propagation. A layer outputs a conditional distribution denoting the extracted
input features at this
layer~\cite{bengio2013representation,bishop1995neural,schmidhuber2015deep}. And
with the extracted features getting more abstract at deeper layers, the ending
layer returns the probability that the input belongs to each label.
Sequential data, such as text, is processed by recurrent models (e.g., LSTM) in
a similar way.

\noindent \textbf{Generative Models.}~Another popular paradigm is the deep
generative model, which estimates the joint probability distribution $P(X,Y)$
of data $X$ and label $Y$. The most common models are VAE~\cite{kingma2013auto} and
GAN~\cite{goodfellow2014generative}. Generative models are primarily adopted
to synthesize realistic data. To do so, both VAE and GAN project the data
distribution to a simple distribution (e.g., normal distribution). To generate
realistic outputs, random inputs sampled from the pre-defined distribution
are fed into deep generative models.

\noindent \textbf{Visualizing DNN Outputs.}~We empirically assess and visualize
the distribution of intermediate neurons' outputs. We prepare a DNN (denoted as
\texttt{Net}; see implementation in~\cite{snapshot}) with two neurons $n_1$ and
$n_2$ (thus the dimensions are two) in
the last layer and extend \texttt{Net} into 1) a VAE, 2) a class-conditional
VAE (C-VAE), and 3) a classifier. The three models are trained on the MNIST
dataset~\cite{deng2012mnist} for 1) random image generation, 2) labeled image
(e.g., ``2'') generation, and 3) image classification. In short, these models
inherit the same structure from \texttt{Net} yet perform different tasks by
using different trailing layers.

\begin{figure}[!ht]
  \centering
  \vspace{-10pt}
  \includegraphics[width=0.9\linewidth]{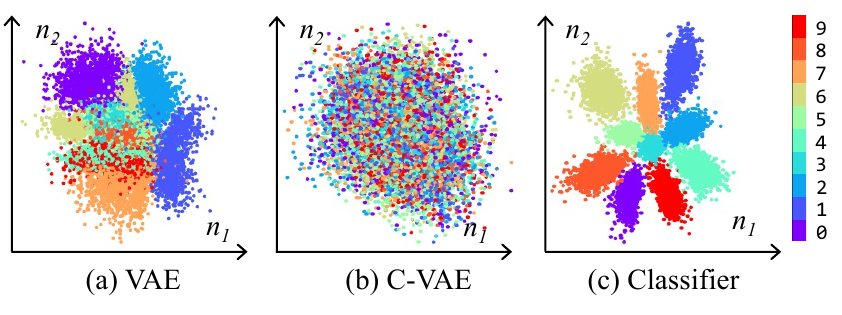}
  \vspace{-5pt}
  \caption{Visualization of $n_1$ and $n_2$'s outputs in three models extended
    from \texttt{Net}. X- and y-axis are outputs of $n_1$ and $n_2$.}
  \label{fig:visual}
  \vspace{-5pt}
\end{figure}

\F~\ref{fig:visual} shows the outputs of $n_1$ and $n_2$ in these models. Each
dot denotes one output of $n_1$ and $n_2$, corresponds to one input. Colors
represent labels of the inputs (e.g., \textcolor{red}{red} represents ``9''). In
\F~\hyperref[fig:visual]{1(a)} and \F~\hyperref[fig:visual]{1(c)}, neuron
outputs are jointly distributed corresponding to different labels, whereas in
\F~\hyperref[fig:visual]{1(b)}, outputs are not split.
Also, outputs gather as regions of varied density and shapes.
\F~\ref{fig:visual} demonstrates that, models sharing the same architecture,
inputs, and output range (e.g., \F~\hyperref[fig:visual]{1(a)}
\F~\hyperref[fig:visual]{1(b)}), may have distinct behaviors, which can
\textit{hardly be assessed using range or distance of neuron outputs by existing
criteria}.

This study captures DNN behaviors by examining the continuity of neuron outputs.
It also considers neuron entanglement, \textit{where neurons in a layer are correlated
and jointly form distributions} that reflect how DNNs (of different tasks)
understand inputs distinctly. 

\begin{figure*}[!ht]
  \centering
  \includegraphics[width=0.85\linewidth]{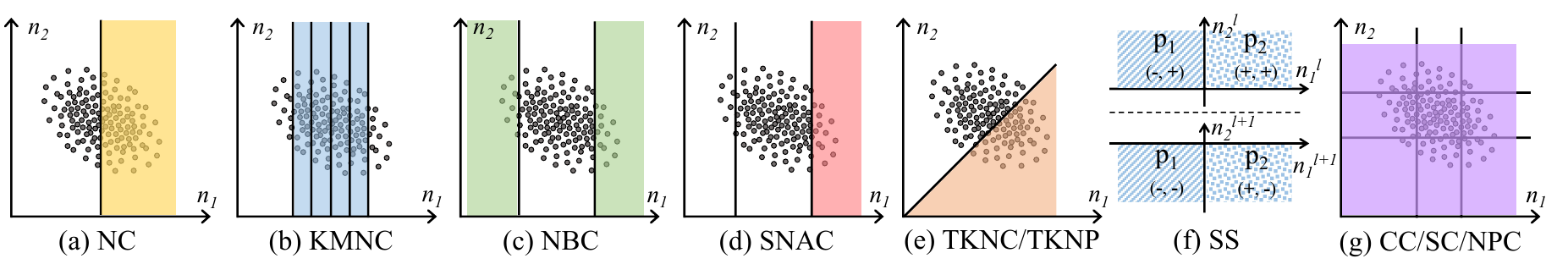}
  \vspace*{-10pt}
  \caption{A schematic view of prior DNN coverage criteria. Aligned to
  \F~\ref{fig:visual}, two axes denote two neurons in a layer.}
  \vspace*{-10pt}
  \label{fig:coverage}
\end{figure*}

\noindent \textbf{NeuraL Coverage.}~Distributions formed by entangled neurons are
important indicators of DNN behaviors. Given that ``neural'', holistically,
correlates to distributions of neuron outputs when understanding input features, this
work proposes \underline{N}eura\underline{L} \underline{C}overage, \tool, a new
criterion that captures properties in distributions formed by neuron outputs. The term
``neural,'' as an adjective, better reflects that neuron outputs are processed in the
continuous form and ``L'' also indicates each \underline{L}ayer is considered as
the minimal computing unit.

\section{Motivation and Related Works}
\label{sec:motivation}

This section, by 1) comparing traditional software and DNN, 2) demonstrating
limitations of prior criteria, and 3) incorporating practical considerations,
proposes \textbf{eight requirements} for DNN coverage criteria that motivates
\tool.

\subsection{DNN Testing and Software Testing}
\label{subsec:comparison}

\noindent \textbf{Discrete vs.~Continuous States.}~Software coverage is often
measured by counting \#statements or \#paths executed by an
input. Software testing aims to maximize code coverage, as high coverage
suggests comprehensive testing. Inspired by software testing, previous DNN
testing often converts a neuron's \textit{continuous}
outputs~\cite{pei2017deepxplore,ma2018deepgauge}, or distance between neuron
output traces~\cite{kim2019guiding,kim2020reducing}, into discrete
states. The coverage is denoted as the \#activated states of a test input.
As shown in \F~\ref{fig:visual}, discrete states are unexpressive for describing
neuron behavior. Instead, we advocate that a DNN coverage criterion should
\CircleOne\textit{precisely describe the continuous output of a neuron}.

\noindent \textbf{Statement vs.~Correlation.}~It is inaccurate to compare
neurons to software statements, because each neuron is \textit{not} an
independent computing unit. In general, an ``unactivated'' neuron
(defined by~\cite{pei2017deepxplore}) can still affect the 
DNN activities and final predictions.
In contrast, software statements may not be covered by an input.
Given that said, \F~\ref{fig:visual} shows that neuron outputs are often
correlated (e.g., increasing simultaneously). Neuron correlation is, to some
extent, analogous to program dependency, e.g., statements within an \texttt{if} block
execute together. Rather than studying individual neurons, it is preferable
(and more efficient) to measure DNN coverage by \CircleTwo\textit{characterizing
correlations of neurons}.

\noindent \textbf{Execution vs.~Estimation.}~Software executes its input on an
execution path. DNNs, as described in \S~\ref{sec:dnn}, use a series of 
layers to approximate certain distributions for an input. Different DNN inputs
produce different distributions, as seen in
\F~\ref{fig:visual}. Hence, we advocate to define the coverage 
criterion over approximated distributions.
While it's difficult to precisely describe the approximated
distribution~\cite{rasmussen1999infinite,reynolds2009gaussian,xuan2001algorithms},
certain informative properties of a distribution can be captured. Also, since
coverage criteria are used to assess test suite quality and guide input
mutations, knowing the exact distribution may not be required (see our empirical
results in \S~\ref{sec:evaluation}). In brief, we advocate that the DNN coverage
criterion should
\CircleThree\textit{analyze how outputs of neurons in a layer are distributed}.

\subsection{Existing Neuron Coverage and Their Limitations}
\label{subsec:nc}

We illustrate existing DNN coverage criteria in \F~\ref{fig:coverage} using two
neurons to ease reading. In general, prior coverage criteria
leverages discretization to 
decide if a state is ``covered.''

\noindent \textbf{Major Behavior of One Neuron.}~Neuron Coverage
(NC)~\cite{pei2017deepxplore}, as the first criterion, argues that the
\textit{major behavior} of a neuron is mostly reflected when its outputs are
above zero. As in \F~\hyperref[fig:coverage]{2(a)}, NC deems a neuron as
activated if its output is above threshold $T$ ($T$ is zero or a small value
above).
Following, three criteria were proposed~\cite{ma2018deepgauge}, namely
K-Multisection Neuron Coverage (KMNC), Neuron Boundary Coverage (NBC) and Strong
Neuron Activation Coverage (SNAC). We display KMNC, NBC and SNAC in
\F~\hyperref[fig:coverage]{2(b)}, \F~\hyperref[fig:coverage]{2(c)} and
\F~\hyperref[fig:coverage]{2(d)}. All three criteria first run training data to
scope the output range of each neuron. Then, KMNC splits each range into $K$
segments and regards coverage as \#segments covered by neuron outputs. NBC
denotes coverage as \#neuron outputs lying outside the range, while SNAC counts
coverage as \#neurons whose outputs are larger than the upper bound of its
normal value range. Their intuition is that the major behaviors of neurons are
denoted as outputs that lie in the range decided by training data.

\noindent\underline{Limitations:}~Considering the \textcolor{blue}{blue} region in \F~\hyperref[fig:visual]{1(c)}, it is easy to see
that when feeding inputs of label ``1'', NC over those two neurons will
saturate. This indicates that criteria, which simply discretize neuron outputs,
cannot properly describe the \textit{diversity} of test suites (e.g., inputs of
different labels). Moreover, since these criteria do not consider neuron entanglement,
they should not change if the \textcolor{orange}{orange distribution} of label
``7'' in \F~\hyperref[fig:visual]{1(c)} is rotated for 90 degrees. That is, these
criteria might hardly reflect distinctly changed DNN behaviors.

Following \CircleOne, we argue that a neuron's \textit{major} behavior can
hardly be defined using coarse-grained thresholds. Instead, given that the
high-density regions formed by neuron outputs in a layer (e.g., the ``dots'' on
\F~\ref{fig:coverage}'s background) are distinct on \textit{shapes}, we deem the
major behavior as the shapes which reflect the majority of neuron activates.
Corner-case behaviors are therefore regarded as neuron outputs
lying in low-density regions. In sum, we argue that \CircleFour\textit{DNN
coverage should consider density of neuron output distributions}. Measuring 
density considers both major and corner-case behaviors of DNNs, which is
desirable in DNN testing.

\noindent \textbf{Discretizing Outputs of Multiple Neurons.}~Two layer-wise
criteria, Top-K Neuron Coverage (TKNC) and Top-K Neuron Patterns (TKNP), were
also proposed in~\cite{ma2018deepgauge}. Both criteria denote a neuron as
activated if its outputs are ranked at top-$K$ among neurons in a layer.
TKNC counts \#neurons once been activated, whereas TKNP records \#patterns formed
by activated neurons. As shown in \F~\hyperref[fig:coverage]{2(e)} where there
are two neurons, a neuron is activated if its output is greater than another
one (i.e., $K = 1$ and $n_1 > n_2$). Sign-Sign coverage (SS)~\cite{sun2018testing} captures the sign change of
layer-wise adjacent neurons. A neuron is activated if 1) its outputs, and
outputs from its adjacent neuron, have different signs for any pair of input
$\langle p_1, p_2 \rangle$ (i.e., $n_1^{l}$ and $n_1^{l+1}$ in
\F~\hyperref[fig:coverage]{2(f)}) and simultaneously, 2) outputs from all other
neurons in the same layer have an identical sign for $\langle p_1, p_2 \rangle$
(i.e., $n_2^{l}$ and $n_2^{l+1}$ in \F~\hyperref[fig:coverage]{2(f)}).

\noindent \underline{Limitations:}~Despite considering all neurons in a layer,
TKNC and TKNP still treat one neuron as a minimal computing unit
and are \textit{incapable} of analyzing the output distribution.
They are easily saturated w.r.t.~inputs of one or a few classes
(see \F~\ref{fig:visual} and \F~\hyperref[fig:coverage]{2(e)}), failing to
assessing the diversity of a test suite. SS is
specifically designed for ReLU-DNNs (i.e., DNNs with ReLU as the non-linear
function). However, since most of the neuron outputs are nearly zero-centered
due to input and layer normalization in modern
DNNs~\cite{ioffe2015batch,sola1997importance}, SS can be less useful for many
of them. Also, while SS considers the relation of adjacent layers, since each
neuron is still a separate unit in SS, the relation of layers is modeled at the
neuron-level.

As noted in \S~\ref{sec:dnn}, the DNN hierarchy is formed by input feature
re-use (e.g., processing shape at shallow layers and then texture at deep
layers~\cite{geirhos2018imagenettrained,zeiler2014visualizing}). The hierarchy
is generally encoded into layer outputs. \tool, by capturing the distribution of
layer outputs, is more desirable compared to SS.

\noindent\textbf{Discretizing Distances over Traces of
Neurons.}~Cluster-based Coverage (CC) assesses distances between traces of
neuron outputs~\cite{odena2018tensorfuzz}. CC counts \#clusters, where two
clusters are formed if, the Euclidean distance of outputs from all neurons
within a layer, is larger than a threshold. Given that more clusters are formed
if neuron outputs get spread, CC can be regarded as counting \#regions covered by
several neuron outputs; as illustrated \F~\hyperref[fig:coverage]{2(g)}.

Surprise adequacy (SA) metrics, namely, Likelihood SA (LSA), Distance-ratio SA
(DSA) and Mahalanobis Distance SA (MDSA), assess the distance between the neuron
output trace of an input and neuron output traces of all training data to decide
how ``surprise'' the new input is~\cite{kim2019guiding,kim2020reducing}. LSA and
DSA require iterating over all traces of neuron outputs collected from training
data. MDSA reduces the computation cost by caching class-conditional mean and
covariance matrices estimated on training data. Surprise Coverage (SC) derived
via SAs discretizes the range of SA values into buckets, and coverage is denoted
as the ratio of covered buckets.

Neuron Path Coverage (NPC)~\cite{xie2022npc} is based on explainable AI
techniques~\cite{bach2015pixel}: a path is constituted by critical neurons
(identified via DNN gradients) from different layers. Similar to SCs, the coverage
is calculated via discretized buckets of distance values over different paths.
Holistically, since more spreading neuron outputs induce higher coverage values
for SCs and NPC, they can also be expressed using \F~\hyperref[fig:coverage]{2(g)}.

\noindent\underline{Limitations:}~CC is more desirable than prior criteria
given it is more comprehensive to overlay the output distribution.
Nevertheless, CC could not capture correlations among neurons. SCs/NPC have similar limitations.
In addition, unlike existing gray-box coverage criteria, NPC requires white-box access
(e.g., DNN weights, back-propagation gradients) to the tested DNN, which
limits its application scope.

\subsection{Practical Considerations}
\label{subsec:practical}

\noindent \textbf{Incorporating Training Data.}~The training of a DNN
involves a ``test-and-fix'' procedure: the DNN is tuned to reduce its
errors and fit better on training data. This way, training data generally
reflects how DNN works inside. We argue that \CircleFive\textit{DNN coverage
should support incorporating prior-knowledge extracted from training data}.
Existing criteria use training data to scope neuron output
ranges~\cite{ma2018deepgauge}. Nevertheless, for some criteria, e.g., TKNC/TKNP,
their hyper-parameter $K$ depends on model structures instead of training data.
\tool\ can use training data to estimate distributions of neuron outputs before
being used in online testing, thereby satisfying \CircleFive.

\noindent\textbf{Matrix-Form Computation.}~Coverage criteria are often
used in costly DNN testing. According to the released code of existing criteria,
we notice that they are mostly written in a \textit{loop-fashion} which is slow.
We find that testings guided by these criteria spend most time (over
$80\%$) on coverage calculation. We re-implemented these criteria in a
\textit{matrix-fashion}, i.e., the calculation is performed as a sequence of
matrix operations. Given that modern DL frameworks (e.g., PyTorch) can optimize
matrix operations and also use hardware
optimizations~\cite{chen2018tvm,rotem2018glow,ma2020rammer}, the calculation is
accelerated over $100$ times; see our codebase at~\cite{snapshot}. Therefore,
we advocate that \CircleSix\textit{a coverage criterion should support matrix-form
computation to be optimizable by modern DL frameworks and hardware}.

\noindent\textbf{Incremental Update.}~DNN coverage is also typically used to guide test input mutations. Since inputs
are gradually fed, we point out that \CircleSeven\textit{a coverage should
feature efficient incremental update}. Ideally, when processing a new input, the
cost for updating coverage should be a small constant. Since CC iterates over
existing clusters and \#clusters increases when more data are produced, it fails
to fulfill \CircleSeven. TKNP suffers from similar drawbacks. SS requires
extensively comparing each neuron output with others and is thus inefficient for
incremental update. Similarly, for each new input, LSC and DSC iterate over all
neuron output traces of training data, resulting in a computing cost of the
(\#training data) $\times$ (\#neurons) magnitude. Compared with other criteria
(except for CC, TKNP, SS) which have a fixed small cost with a magnitude of
(\#neurons) for each incremental update, LSC and DSC fail to satisfy
\CircleSeven.

\noindent\textbf{User-Friendliness.}~Most existing criteria rely on hyper-parameters. When using existing criteria,
we find that coverage results are usually highly sensitive to hyper-parameters
(see \S~\ref{subsubsec:diversity}),
whereas existing works hardly provide guidance to decide hyper-parameters under
different scenarios. Viewing that deciding hyper-parameters likely requires
expertise (in both software testing and deep learning), we advocate that
\CircleEight\textit{a DNN coverage is desirable to be hyper-parameter free.}

\begin{table}[t]
  \vspace{-5pt}
  \caption{Benchmarking coverage criteria for DNN testing. \CBrush, \TriUp,
    \XBrush\ denote support, partially support, and not support.
    \protect\Grid{gray} denotes gray-box (only needs neuron outputs)
    and \protect\Grid{white} denotes white-box that requires DNN weights
    and gradients.}
   \vspace{-5pt}
  \label{tab:nc}
  \centering
\resizebox{0.85\linewidth}{!}{
  \begin{tabular}{l|c|c|c|c|c|c|c|c}
    \hline
               & \CircleOne & \CircleTwo & \CircleThree & \CircleFour & \CircleFive & \CircleSix & \CircleSeven & \CircleEight \\
    \hline
    \Grid{gray} NC~\cite{pei2017deepxplore} & \XBrush & \XBrush & \XBrush & \XBrush & \CBrush & \CBrush & \CBrush & \XBrush \\
    \hline
    \Grid{gray} KMNC~\cite{ma2018deepgauge} & \XBrush & \XBrush & \XBrush & \TriUp & \CBrush & \CBrush & \CBrush & \XBrush \\
    \hline
    \Grid{gray} NBC~\cite{ma2018deepgauge} & \XBrush & \XBrush & \XBrush & \TriUp & \CBrush & \CBrush & \CBrush & \CBrush \\
    \hline
    \Grid{gray} SNAC~\cite{ma2018deepgauge} & \XBrush & \XBrush & \XBrush & \TriUp & \CBrush & \CBrush & \CBrush & \CBrush \\
    \hline
    \Grid{gray} TKNC~\cite{ma2018deepgauge} & \XBrush & \TriUp & \XBrush & \XBrush & \XBrush & \CBrush & \CBrush & \XBrush \\
    \hline
    \Grid{gray} TKNP~\cite{ma2018deepgauge} & \XBrush & \TriUp & \XBrush & \XBrush & \XBrush & \CBrush & \XBrush & \XBrush \\
    \hline
    \Grid{gray} SS~\cite{sun2018testing} & \XBrush & \TriUp & \XBrush & \XBrush & \XBrush & \XBrush & \XBrush & \CBrush \\
    \hline
    \Grid{gray} CC~\cite{odena2018tensorfuzz} & \XBrush & \XBrush & \TriUp & \TriUp & \CBrush & \CBrush & \XBrush & \XBrush \\
    \hline
    \Grid{gray} LSC~\cite{kim2019guiding} & \XBrush & \XBrush & \TriUp & \TriUp & \CBrush & \CBrush & \XBrush & \XBrush \\
    \hline
    \Grid{gray} DSC~\cite{kim2019guiding} & \XBrush & \XBrush & \TriUp & \XBrush & \CBrush & \CBrush & \XBrush & \XBrush \\
    \hline
    \Grid{gray} MDSC~\cite{kim2020reducing} & \XBrush & \XBrush & \TriUp & \XBrush & \CBrush & \CBrush & \CBrush & \XBrush \\
    \hline
    \Grid{white} NPC~\cite{xie2022npc} & \XBrush & \XBrush & \TriUp & \TriUp & \CBrush & \CBrush & \CBrush & \XBrush \\
    \hline
    \Grid{gray} NLC & \CBrush & \CBrush & \CBrush & \CBrush & \CBrush & \CBrush & \CBrush & \CBrush \\
    \hline
  \end{tabular}
  }
  \vspace{-15pt}
\end{table}

\subsection{Comparison with Previous Criteria}
\label{subsec:requirement}

\T~\ref{tab:nc} benchmarks
existing DNN coverage criteria (and \tool) w.r.t.~the eight requirements
summarized in this section. Note that TKNC/TKNP consider the relative relation
of neuron outputs and SS considers the sign relation between two neurons. They
are deemed as partially satisfying \CircleTwo. Though CC/SCs/NPC consider a
trace of neurons, how one neuron responds to others can hardly be captured.
They thus fail to satisfy \CircleTwo. Given that the spreading of neuron outputs
is captured, they partially fulfill \CircleThree.
Also, since neuron outputs get concentrated, the neuron output range overlays
the high-density regions. Similarly, the cluster/merged paths in CC/NPC and
likelihood in LSC reflect high and low density regions respectively. We thus
regard KMNC, NBC, SNAC, CC, and LSC as partially satisfying \CircleFour.
In contrast, \tool\ satisfies all eight requirements, as explained in
\S~\ref{sec:design}.

\section{Design of NeuraL Coverage}
\label{sec:design}

\begin{figure}[!ht]
  \centering
  \vspace{-5pt}
  \includegraphics[width=1.0\linewidth]{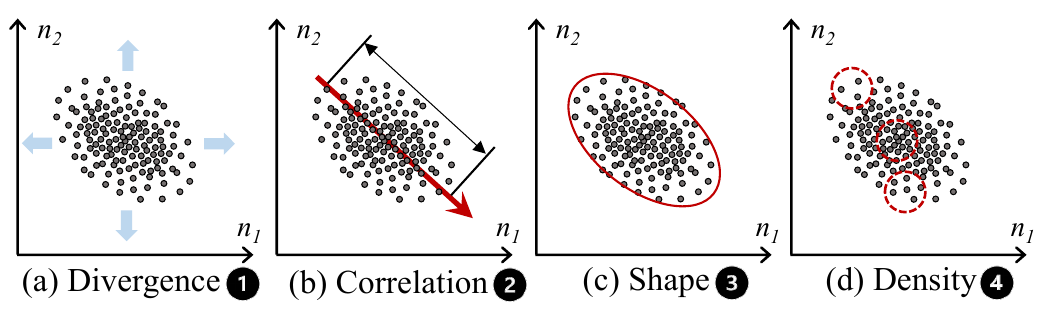}
  \vspace{-20pt}
  \caption{Four properties describing neuron output distributions. We take
    these properties into account to design \tool.}
  \vspace{-5pt}
  \label{fig:tool}
\end{figure}

As noted in \S~\ref{subsec:comparison}, it's generally difficult to precisely
and efficiently describe distributions approximated by
DNNs~\cite{rasmussen1999infinite,reynolds2009gaussian,xuan2001algorithms}.
Instead, as depicted in \F~\ref{fig:tool}, \tool\ captures four key properties of
distributions: divergence, correlation, shape, and density, which corresponds to
the first four criteria in \T~\ref{tab:nc}.
\S~\ref{sec:evaluation} presents empirical results to show that these properties
are sufficient to form a criterion that outperforms all existing works. We now
elaborate on each property.

\noindent\textbf{Measuring Activation in Continuous Space.}~Neuron outputs are
continuous floating-point numbers. Instead of discretizing continuous neuron outputs,
or distances between neuron output traces, \tool\ directly
measures neuron outputs in the continuous space. We start with the case of one
neuron.

\begin{figure}[!ht]
  \centering
  \vspace{-10pt}
  \includegraphics[width=0.95\linewidth]{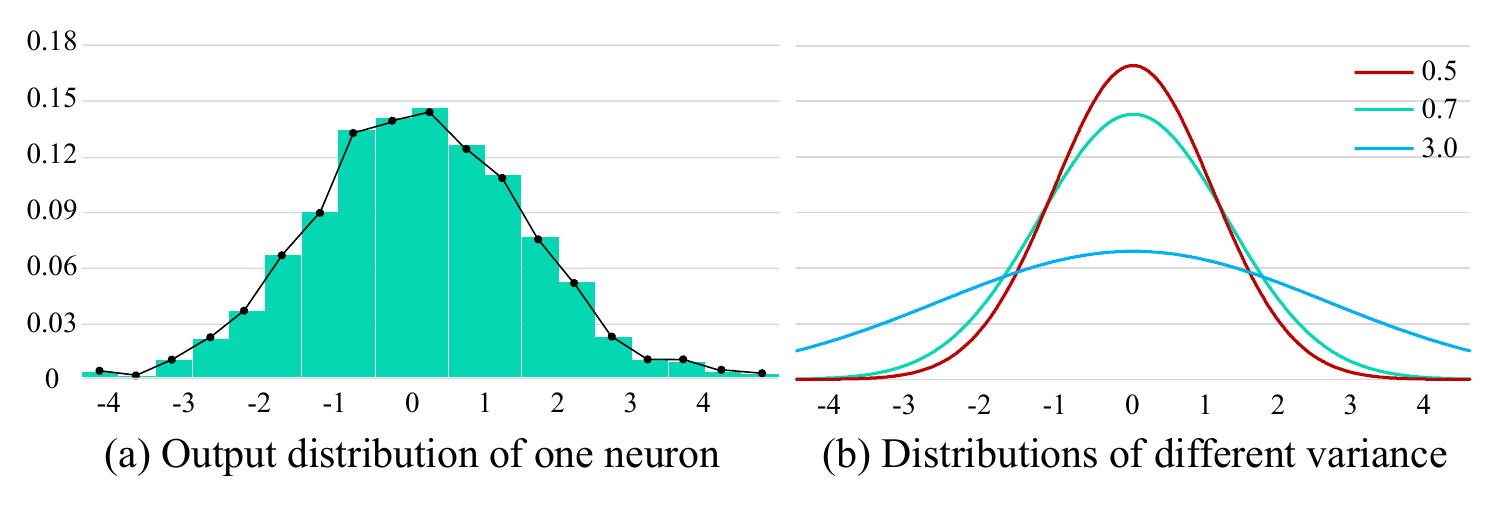}
  \vspace{-8pt}
  \caption{Output distribution of one neuron from ResNet50.}
  \vspace{-10pt}
  \label{fig:diverse}
\end{figure}

We randomly select one neuron from a pre-trained ResNet50 and plot its output
distribution in \F~\hyperref[fig:diverse]{4(a)} using all images in its
training dataset. The x-axis denotes the output value and the y-axis
denotes the ratio of involved inputs. Consistent with \F~\ref{fig:visual},
output values converge to a certain point (zero in this case). \tool\ directly
measures how divergent (i.e., how active) the neuron output is in its
continuous form by calculating the following variance:

\vspace{-5pt}
\begin{equation}
    \label{equ:var}
    \sigma_n^2 = \mathbb{E}[(o - \mathbb{E}[o]) (o - \mathbb{E}[o])],
\end{equation}
\vspace{-10pt}

\noindent where $o$ is the output of a neuron $n$. Overall, variance is widely-adopted
for characterizing the divergence of a collection of samples. \E~\ref{equ:var}
increases when output values spread along the x-axis. In contrast, if most
outputs are zero, the variance will decrease, as in
\F~\hyperref[fig:diverse]{4(b)}. Variance serves as an important descriptor of
neuron output distributions. A test suite is regarded as comprehensive if it
results in a high variance. Similarly, input mutations can be launched towards
the direction that increases variance. Compared with previous criteria, we
require no discretization, thus taking all distinct neuron outputs into account
(in response to \CircleOne). In addition, \E~\ref{equ:var} does not ship with
hyper-parameters (respond to \CircleEight).
It's worth noting that variance is non-monotonic. That is, its value may decrease
given new inputs. To alleviate this problem, we only update it when its value
increases (i.e., $\Delta\sigma_n^2 = \max(0, \Delta\sigma_n^2)$). Generally,
a stable variance, which stays constantly or tends to decrease, indicates that
the behaviors are (nearly) fully explored.

\begin{figure}[!ht]
  \vspace{-5pt}
  \centering \includegraphics[width=0.8\linewidth]{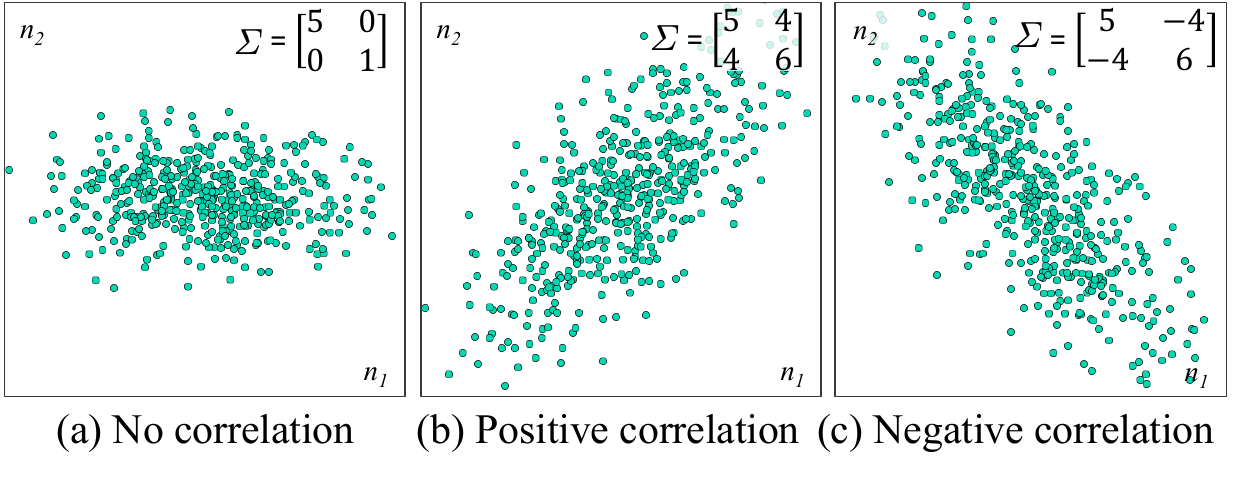}
  \vspace{-5pt}
  \caption{Neurons having various correlations.}
  \vspace{-5pt}
  \label{fig:relation}
\end{figure}

\noindent \textbf{Characterizing Neuron Correlation.}~As noted in
\S~\ref{sec:dnn}, neurons in a layer are entangled and jointly process input
features. As empirically assessed in \F~\ref{fig:visual}, neurons manifest
distinct correlations under different tasks and for inputs of different classes.
For example, in \F~\hyperref[fig:visual]{1(c)}, neuron $n_1$ is positively (e.g.,
class \textcolor{blue}{``1''}) or negatively (e.g., class
\textcolor{red}{``9''}) correlated with neuron $n_2$.

\F~\ref{fig:relation} presents three different correlations of two neurons. Given
that there are more than one neuron, the DNN coverage should simultaneously
consider the divergence of each neuron and correlations among neurons.
\E~\ref{equ:var} uses variance to describe divergence of a neuron's output,
i.e., $\sigma^2_{n_1}$ and $\sigma^2_{n_2}$. Similarly, we use the covariance, a
common metric for joint variability of two variables, to characterize the
correlation between two neurons. Thus, the correlation between the two neurons
is formulated as

\vspace{-10pt}
\begin{equation}
    \label{equ:covar}
    \varsigma_{n_1, n_2} = \mathbb{E}[(n_1 - \mathbb{E}[n_1]) (n_2 - \mathbb{E}[n_2])] = \varsigma_{n_2, n_1}.
\end{equation}
\vspace{-5pt}

\noindent $\varsigma_{n_1, n_2} = 0$ implies that $n_1$ and $n_2$ are not
correlated, whereas $\varsigma_{n_1, n_2} > 0$ denotes a positive correlation
between $n_1$ and $n_2$, i.e., $n_1$ will increase with $n_2$, and vice verse. A
higher $\varsigma_{n_1, n_2}$ denotes a stronger correlation. The same holds
when $\varsigma_{n_1, n_2} < 0$. Taking \E~\ref{equ:var} and \E~\ref{equ:covar}
into account, we simultaneously represent the divergence of each neuron and
correlations of neurons in a unified form as

\vspace{-5pt}
\begin{equation}
    \label{equ:matrix}
    \varSigma = 
    \begin{bmatrix}
        \sigma^2_{n_1} & \varsigma_{n_1, n_2} \\
        \varsigma_{n_2, n_1} & \sigma^2_{n_2}
    \end{bmatrix},
\end{equation}
\vspace{-5pt}

\noindent where $\varSigma$ is the covariance matrix of $n_1$ and $n_2$. As we
show in \F~\ref{fig:relation}, the value of $\varsigma_{n_1, n_2}$ also
represents the divergence along the ``correlation direction'' of two neurons.
Thus, we use

\vspace{-5pt}
\begin{equation}
    \label{equ:coverage}
    \frac{1}{m \times m} \|\varSigma\|_1 = \frac{1}{m \times m}
    \sum\nolimits_{j=1}^{m}\left( \sum\nolimits_{i=1}^{m}|\varsigma_{n_i, n_j}| \right),
\end{equation}
\vspace{-5pt}

\noindent where $m$ is \#neurons and $\varsigma_{n_i, n_i} = \sigma^2_{n_i}$, to
represent the DNN coverage. The size of $\varSigma$ is $m \times m$ and
$\frac{1}{m \times m}$ works as the normalization term to eliminate the effect
of \#neurons to the coverage values.
In the case where more than two neurons are correlated, e.g., $n_i$, $n_j$ and
$n_k$, $\varSigma$ will characterize the correlation via $\varsigma_{n_i,
  n_j}$, $\varsigma_{n_i, n_k}$ and $\varsigma_{n_j, n_k}$. Hence,
\E~\ref{equ:coverage} facilitates addressing requirements \CircleOne\ and \CircleTwo.
Note that $\varSigma$ can be initialized using the correlation term
$\varsigma_{n_i, n_i}$ decided by training data.
Also, $\varSigma$ can be further refined in a class conditional manner by extending
the size to $c \times m \times m$ where $c$ is \#classes. It thus will be calculated
by first being indexed using the class labels. In both two cases, \tool\ fulfills the
requirement of \CircleFive. Further, given a DNN contains multiple layers,
the coverage of DNN is computed as

\vspace{-5pt}
\begin{equation}
  \label{equ:nlc}
  \tool\ = \sum_{l} \frac{1}{m^l \times m^l} \|{\varSigma}^l\|_1,
\end{equation}
\vspace{-5pt}

\noindent where $l$ is the layer index. A higher \tool\ indicates a more diverse
test suite. Test input mutations, accordingly, are expected to accumulate towards
directions that maximize \tool.

\noindent\textbf{Capturing Distribution Shape.}~Unlike previous
works where each neuron is considered separately, we take neurons from the same
layer as a group. As in \F~\ref{fig:visual}, the output distribution of neurons
in the same layer is mostly distinct under different tasks. Moreover, for the
separated regions (in different colors) under certain tasks (e.g.,
classification), they may also have different shapes characterizing various DNN
behaviors. Hence, we capture a major behavior by describing the shape of a
distribution. We now explain why $\varSigma$ in \E~\ref{equ:matrix} captures
a distribution's shape.

\begin{figure}[!ht]
    \centering
    \vspace{-5pt}
    \includegraphics[width=1.00\linewidth]{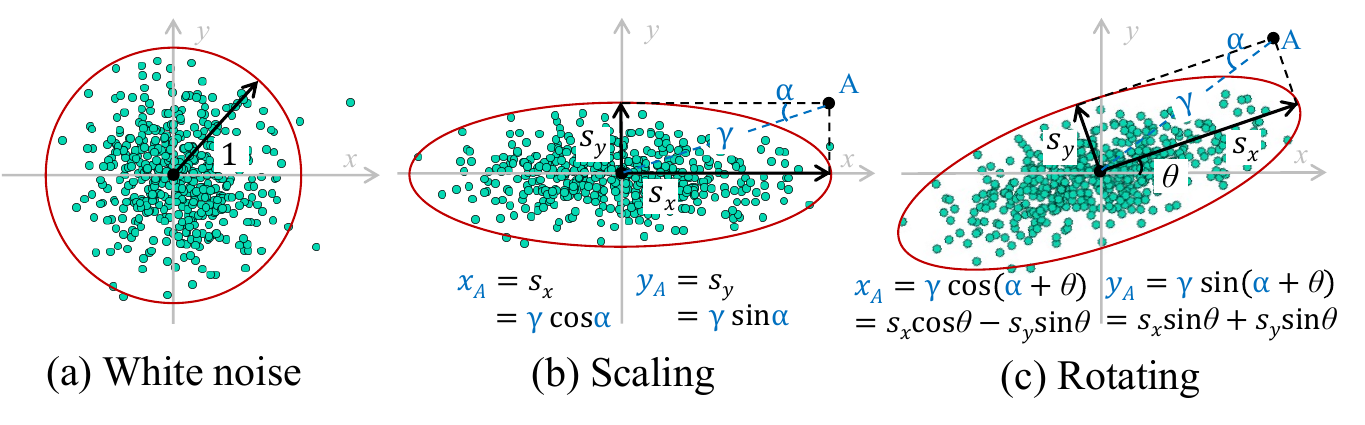}
    \vspace{-15pt}
    \caption{Transforming white noise to a desired distribution.}
    \label{fig:shape}
    \vspace{-5pt}
\end{figure}

To ease understanding, we first show how to transform a white noise (whose
covariance matrix is degenerated into an identity matrix $I$) to the
distribution formed by outputs from a group of neurons.
\F~\hyperref[fig:shape]{6(a)} shows a white noise $\mathcal{N}$ and
\F~\hyperref[fig:shape]{6(c)} presents the output distribution of two neurons
$x$ and $y$. The transformation is thus decomposed into two elementary operators
--- scaling and rotating. The scaling operator spreads two neuron outputs and
rotating introduces correlations between $x$ and $y$. Rotating and scaling are
defined by rotation matrix and scaling matrix as

\vspace{-5pt}
\begin{equation}
    \label{equ:transform}
    \mathcal{R}_{2 \times 2} = 
    \begin{bmatrix}
        \cos\theta & -\sin\theta \\
        \sin\theta & \cos\theta \\
    \end{bmatrix},
    \mathcal{S}_{2 \times 2} = 
    \begin{bmatrix}
        s_x & 0 \\
        0 & s_y \\
    \end{bmatrix},
\end{equation}
\vspace{-5pt}

\noindent respectively. Thus, the output distribution can be represented as
$(\mathcal{R}\mathcal{S})\mathcal{N}$. Following \E~\ref{equ:var} and
\E~\ref{equ:covar}, we have

\vspace{-5pt}
\begin{equation}
  {\small
    \begin{aligned}
    \label{equ:prove}
    \varSigma &= \mathbb{E}[(\mathcal{R}\mathcal{S}\mathcal{N} - \mathbb{E}[\mathcal{R}\mathcal{S}\mathcal{N}])
                             (\mathcal{R}\mathcal{S}\mathcal{N} - \mathbb{E}[\mathcal{R}\mathcal{S}\mathcal{N}])^{\top}] \\
               &= (\mathcal{R}\mathcal{S})
                  \mathbb{E}[(\mathcal{N} - \mathbb{E}[\mathcal{N}]) (\mathcal{N} - \mathbb{E}[\mathcal{N}])^{\top}]
                  (\mathcal{R}\mathcal{S})^{\top} \\
               &= (\mathcal{R}\mathcal{S})I(\mathcal{R}\mathcal{S})^{\top} = (\mathcal{R}\mathcal{S}) (\mathcal{R}\mathcal{S})^{\top}, \\
    \end{aligned}
  }
\end{equation}
\vspace{-5pt}

\noindent which indicates that covariance matrix $\varSigma$ in
\E~\ref{equ:matrix} encodes transformations from a white noise to a desired
distribution. Therefore, we demonstrate that $\varSigma$ captures
the shape (i.e., how it is distributed) of the output distribution formed by outputs of neurons in a
layer, satisfying \CircleThree.

\begin{figure}[!ht]
    \centering
    \vspace{-10pt}
    \includegraphics[width=0.85\linewidth]{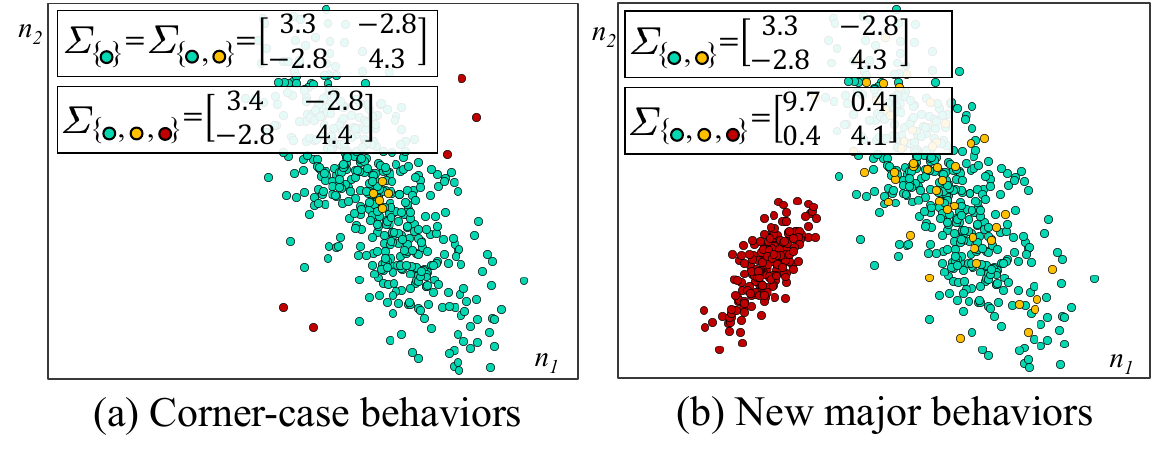}
    \vspace{-5pt}
    \caption{Responses to density changes.}
    \vspace{-5pt}
    \label{fig:density}
\end{figure}

\noindent\textbf{Responding to Density Change.}~Neuron outputs generally
concentrate to certain points to form clusters, and the density varies in the
output space. We first elaborate on what may introduce a local density change.
Recall that we deem corner-case behaviors as neuron outputs lying at the low-density
regions (e.g., boundary of one cluster), which are likely to trigger DNN defects.
Given that these outliers typically have much lower local-density, their existence
will lead to density change. 
Consider \F~\hyperref[fig:density]{7(a)} where the DNN outputs in
\textcolor{pptred3}{red} are deemed as outliers. Nevertheless, since they are
still in the output ranges formed by all \textcolor{pptgreen3}{green dots},
SNAC and NBC will not treat them as corner case behaviors. In fact, the new
\textcolor{pptred3}{red} outputs will \textit{not} induce coverage increase for
all previous criteria except CC and SCs, because CC and SCs capture how neuron
outputs spread in the whole space. However, they require properly selected
distance thresholds, which are hyper-parameters configured by users.

Also, for certain cases where neuron outputs are separated (e.g., the classifier
in \F~\hyperref[fig:visual]{1(c)}), the appearance of a new major behavior 
(e.g., images get classified as a new class) also leads to density change.
As in \F~\hyperref[fig:density]{7(b)}, the new cluster of
\textcolor{pptred3}{red dots} introduces a new major behavior, whereas NC
possibly has no response since their values are lower than the threshold.

We show that \tool\ responds to density change. For
\F~\hyperref[fig:density]{7(a)}, the $||\varSigma||_1$ of existing
\textcolor{pptgreen3}{green dots} is $13.2$. It will be updated as $13.4$ given
the new \textcolor{pptred3}{red dots}, which indicates that NLC responds.
In contrast, since the new \textcolor{pptyellow3}{yellow dots} stay close to
the majority of existing outputs, NLC has no response.
It's worth mentioning that, despite the density changes only slightly, NLC has
\textit{no} resolution loss given its continuous form.
Meanwhile, for \F~\hyperref[fig:density]{7(b)}, the $||\varSigma||_1$ will
change from $13.2$ to $14.6$ when the cluster formed by \textcolor{pptred3}{red
  dots} appears. We thus state that our criterion can respond to density change
which is possibly introduced by corner-case behaviors and new major behaviors.
This way, \tool\ satisfies \CircleFour.

\noindent \underline{Clarification on Covariance.}~Both MDSC and NLC rely on covariance.
We, however, clarify that the two ``covariances'' imply distinct concepts.
\tool\ characterizes how fresh inputs change the neuron correlation or introduce
new distributions, using an actively-updated covariance. \tool\ incrementally
updates covariance given new inputs and rollbacks the value if the new input is
not desired. In contrast, the covariance in MDSC is a pre-computed constant
that represents the distribution of historical neuron outputs, based on which
the Mahalanobis distance can be computed from.

\noindent\textbf{Implementation Considerations.}~As discussed above, it
is easy to see that \tool\ supports matrix-form computing and has no
hyper-parameter. Thus, \tool\ satisfies \CircleSix\ and \CircleEight.
We now clarify that \tool\ supports incremental
computing (requirement of \CircleSeven). We also show that \tool\ supports batch
computation, which is also vital as DNNs generally take a batch of inputs and
the batch size varies from $1$ to hundreds.
Let existing data be $\mathbb{B}$, given a batch of new data $B$, the
covariance matrix $\varSigma$ (\E~\ref{equ:coverage}) of each layer can be
calculated as

\vspace{-5pt}
\begin{equation}
    \begin{aligned}
    \label{equ:incremental}
    \varSigma &= \frac{|\mathbb{B}| \times |B| \times (\mu_{\mathbb{B}} - \mu_{B})(\mu_{\mathbb{B}} - \mu_{B})^{\top}}
                {(|\mathbb{B}| + |B|)^2} \\
             &+ \frac{|\mathbb{B}| \times \varSigma_{\mathbb{B}} + |B| \times \varSigma_{B}}
               {|\mathbb{B}| + |B|},
    \end{aligned}
\end{equation}
\vspace{-5pt}

\noindent where $\mu$ is the mean of a collection of data. Clearly,
\E~\ref{equ:incremental} does \textit{not} iterate over existing
data and the computing cost is in a magnitude of (\#neurons);
\tool\ thus satisfies \CircleSeven.

\section{Implementation and Configurations}
\label{sec:implementation}

\tool\ is implemented using PyTorch (ver. 1.9.0) with roughly 2,300 LOC. All
experiments are launched on one Intel Xeon CPU E5-2683 with 256GB RAM and one
Nvidia GeForce RTX 2080 GPU. Below we briefly introduce settings of prior
criteria compared with \tool. Their implementations are released in~\cite{snapshot}.

\noindent \textbf{Optimization and Hyper-parameters.}~We perform optimizations
on our end for all prior criteria examined in \S~\ref{sec:evaluation} in order
to report their best results. In original SCs, coverage is calculated as the
ratio of covered buckets. More specifically, given the maximal SA value $U$ and
maximal bucket count $m$, SCs first discretize $U$ as $m$ buckets of size
$\frac{U}{m}$, then, they calculate coverage as the ratio of covered bucket
(by test suites) over the total $m$ buckets. Nevertheless, our preliminary study
shows that it is \textit{uneasy} to tune these two parameters, $U$ and $m$, separately.
Thus, we use the bucket size $T=\frac{U}{m}$, which represents $U$ and $m$ with
only one variable $T$, as the hyper-parameter for tuning.
Accordingly, we report the \textit{number} of covered buckets in
\S~\ref{sec:evaluation}, which is equivalent to the original ratio of covered
buckets. Suppose our reported coverage value is $|S|$, which denotes the number
of elements in $S$ and $S$ is the set of indexes of covered buckets. To convert
it into the ratio used in original SCs, users need to first choose a maximal SA
value $U$, and obtain $S' = \{s | T s \leq U, s \in S\}$. The ratio of
covered buckets can be computed as $\frac{T|S'|}{U}\%$.

Additionally, as described in \S~\ref{subsec:nc}, we re-implement the prior
criteria's calculation in a matrix form to boost their computation efficiency.
We denote the hyper-parameters of KMNC, TKNC, and TKNP as $K$ in
\S~\ref{sec:evaluation}. In other criteria, thresholds are denoted as $T$.

\noindent \textbf{SS and NPC.}~SS is specifically designed for ReLU-DNN, whereas
modern DNNs have various non-linear functions which are not supported by SS. Also,
SS does \textit{not} support matrix-form and incremental computing. Our
preliminary study shows that it is computational infeasible due to
the large number of neurons in modern DNN.
NPC requires white-box DNN, which is not always practical in testing scenarios.
In addition, its building block, the layer-wise propagation~\cite{bach2015pixel},
is only applicable to discriminative models.
We thus omit to compare with them.

\section{Evaluation}
\label{sec:evaluation}

%
%

\subsection{Assessing Test Suite Quality}
\label{subsec:rq1}

We first compare \tool\ with existing criteria on its effectiveness to assess
the 1) diversity and 2) fault-revealing capability of test suites.

\noindent \textbf{Settings \& DNNs \& Datasets.}~We evaluate \tool\ on both discriminative and
generative models. The involved models and datasets are listed in
\T~\ref{tab:model-rq1} and \T~\ref{tab:data-rq1}. For discriminative models, we
choose VGG16, ResNet50, and MobileNetV2, which represent standard large-scale
DNNs with sequential and non-sequential topological structures and popular
models on mobile platforms. The three models are trained on two
datasets of different scales, namely, CIFAR10~\cite{krizhevsky2009cifar} and
ImageNet~\cite{deng2009imagenet}. Models trained on CIFAR10 have over $92\%$
test accuracy and those trained on ImageNet are provided by PyTorch. CIFAR10
is composed of $32 \times 32$ images labeled as ten classes. ImageNet contains
images of $1,000$ classes and the average image size is $469 \times 387$.
We also evaluate \tool\ on the text model, i.e., a LSTM~\cite{hochreiter1997lstm}
trained on the IMDB dataset~\cite{maas2011learning} for sentiment
analysis~\cite{feldman2013techniques}. Given that each layer in the LSTM is
unrolled to process sequential inputs, the unrolled layers are considered
separately for computing coverage~\cite{tian2018deeptest}.
For generative models, we choose the SOTA model BigGAN. The BigGAN is also
trained on CIFAR10 and ImageNet.

\noindent \textbf{Forming Ground Truth.}~In practice, it is infeasible to
compare coverage values to the ground truth (e.g., a test suite covers 70\% of
all ``diversity'' which is inaccessible due to the infinite number of possible
inputs in the wild). Besides, the coverage value is hardly interpretable (see
discussion in \S~\ref{sec:discussion}). Consequently, the expected magnitude of
the coverage induced by a test suite is unknown. To overcome these hurdles, we
construct cases where the \textit{relative order} of coverage values induced by
different test suites is quantifiable (e.g., which test suite results in
\textit{relatively larger} coverage). The relative order forms a reasonable and
measurable ground truth against which coverage values can be compared to
determine their accuracy.

\subsubsection{Diversity of Test Suites}
\label{subsubsec:diversity}

\noindent \textbf{Discriminative Models.}~To assess the diversity of test suites for
discriminative models, we construct two datasets of different diversity using
schemes denoted as $\times 1$ and $\times 10$. For the $\times 1$ scheme, we
randomly choose $100$ images from the test data and add white noise to produce
dummy images (the total number of dummy images in $\texttt{test}_{\times 1}$
equals to the size of all test data). The white noise is truncated within
$[-0.1, 0.1]$ to preserve the appearance of images. Given that text is discrete,
it is infeasible to add white noise. We thus randomly shuffle words in the $100$
selected sentences to produce dummy sentences. The $\times 10$ scheme
follows the same procedure, but the number of the dummy data is $10$ times of
the test data.

\noindent \underline{Ground Truth (\T~\ref{tab:text-rq1} and \T~\ref{tab:disc-rq1}):}~The diversity of these datasets are ranked as
\underline{$\texttt{test} > \texttt{test}_{\times 10} > \texttt{test}_{\times 1}$},
because $\texttt{test}$ contains over 10K diverse and meaningful samples in real-world
datasets, whereas $\texttt{test}_{\times 10}$ and $\texttt{test}_{\times 1}$ are
generated by (slightly) mutating 100 data samples.

\begin{table}[t]
  \caption{DNN Models used in \S~\ref{subsec:rq1}. All image models are trained on CIFAR10 and ImageNet.}
   \vspace{-5pt}
  \label{tab:model-rq1}
  \centering
\resizebox{0.9\linewidth}{!}{
  \begin{tabular}{l|c|c|c}
    \hline
     Model         & \#Neuron & \#Layer  & Remark \\
    \hline
    ResNet50~\cite{he2016resnet}     & 26,570/27,560      & 54           & Non-sequential topology \\
    \hline
    VGG16\_BN*~\cite{simonyan2014very}   & 12,426/13,416      & 16           & Sequential topology \\ 
   \hline
    MobileNet-V2~\cite{howard2017mobilenets}  & 17,066/18,056      & 53           & Mobile devices \\ 
    \hline
    BigGAN~\cite{brock2018biggan} & 6,403/48,627 & 11/41 & Generative model \\
    \hline
    LSTM~\cite{hochreiter1997lstm}  & 32,769 & 129 & Text model \\
    \hline
  \end{tabular}
  }
  \begin{tablenotes}
    \footnotesize
    \item *All image models have batch normalization (BN) layers.
  \end{tablenotes}
  \vspace{-5pt}
\end{table}

\begin{table}[t]
  \caption{Datasets (train/test) used in \S~\ref{subsec:rq1}. Each class has the same number of images.}
  \vspace{-5pt}
  \label{tab:data-rq1}
  \centering
\resizebox{0.8\linewidth}{!}{
  \begin{tabular}{l|c|c|c}
    \hline
     Dataset   & CIFAR10 & ImageNet & IMDB \\
    \hline
     \#Classes & 10      & 1,000      & 2     \\
    \hline
     \#Data    & 50,000/10,000  & 1,000,000/50,000  & 17,500/17,500 \\
    \hline
  \end{tabular}
  }
 \vspace{-10pt}
\end{table}

\begin{table}[t]
  \caption{Coverage achieved by different (text) test suites on LSTM.
  Assessments  matching ground truth are \colorbox{pptgreen3!25}{marked}.}
  \label{tab:text-rq1}
  \centering
\resizebox{0.85\linewidth}{!}{
  \begin{tabular}{c|c|c|c|c}
      \hline
      Criteria                                 & Config.     & IMDB & $\text{IMDB}_{\times 10}$ & $\text{IMDB}_{\times 1}$ \\
      \hline
      \multirow{3}{*}{\shortstack{NC ($\%$)}}  & $T$=$0.25$ & 0.012  & 0.034  & 0.018                          \\
                                                 & $T$=$0.50$ & 0.015  & 0.027  & 0.018                          \\
                                                 & $T$=$0.75$ & 2.191  & 2.554  & 1.001                          \\
      \hline
      \multirow{3}{*}{\shortstack{KMNC ($\%$)}} & $K$=$10$    & 7.68  & 12.11  & 3.69                              \\
                                                 & $K$=$100$   & 1.92  & 5.31  & 2.99                              \\
                                                 & $K$=$1,000$ & 0.05  & 0.15  & 0.08                              \\
      \hline
      NBC ($\%$)                                 & N/A       & 0.12  & 0.13  & 0.04                              \\
      \hline
      SNAC ($\%$)                                & N/A      & 0.11  & 0.03  & 0.03                              \\
      \hline
      \multirow{3}{*}{\shortstack{TKNC ($\%$)}} & $K$=$1$  & 0.28  & 0.67  & 0.19                              \\
                                                 & $K$=$10$ & 0.50  & 0.96  & 0.39                              \\
                                                 & $K$=$50$ & 1.34  & 2.86  & 1.20                              \\
      \hline
      \multirow{3}{*}{\shortstack{TKNP ($\#$)}} & $K$=$1$  & 20,252 & 101,789  & 12,118                             \\
                                                 & $K$=$10$ & 6,515  & 8,602  & 2,224                              \\
                                                 & $K$=$50$ & 11,830  & 20,479 & 4,633                              \\
      \hline
      \multirow{3}{*}{\shortstack{CC ($\#$)}}   & $T$=$10$   & 266  & 283  & 265                              \\
                                                 & $T$=$20$   & 134  & 146  & 137                              \\
                                                 & $T$=$50$   & 57  & 65  & 1                                \\
      \hline
      \multirow{3}{*}{\shortstack{LSC ($\#$)}}   & $T$=$1$   & 306  & 314  & 214                              \\
                                                 & $T$=$10$   & 40  & 60  & 27                              \\
                                                 & $T$=$100$  & 5  & 7   & 7                                \\
      \hline
      \multirow{3}{*}{\shortstack{DSC ($\#$)}}   & $T$=$0.01$   & 176  & 185  & 153                              \\
                                                 & $T$=$0.1$  &  \colorbox{pptgreen3!25}{25}  & \colorbox{pptgreen3!25}{24}  & \colorbox{pptgreen3!25}{18}                             \\
                                                 & $T$=$1$   & 4   & 4  & 4                                \\
      \hline
      \multirow{3}{*}{\shortstack{MDSC ($\#$)}}   & $T$=$1$   & 101  & 229  & 97                              \\
                                                 & $T$=$10$    & 18  & 43  & 22                              \\
                                                 & $T$=$100$   & 4   & 5  & 5                                \\
      \hline
      \textbf{NLC}                               & N/A     & \colorbox{pptgreen3!25}{1279.81}  & \colorbox{pptgreen3!25}{132.99}  & \colorbox{pptgreen3!25}{11.01}                              \\
      \hline
  \end{tabular}
}
\vspace{-10pt}
\end{table}

\begin{table*}[t]
  \caption{Coverage achieved by different (image) test suites on discriminative models.
  \texttt{CR} and \texttt{IN} denote CIFAR10 and ImageNet dataset, respectively.
  Assessments matching ground truth are \colorbox{pptgreen3!25}{marked}.}
  \label{tab:disc-rq1}
  \centering
\resizebox{1.0\linewidth}{!}{
  \begin{tabular}{
    @{\hspace{1pt}}c@{\hspace{1pt}}|
    @{\hspace{1pt}}c@{\hspace{1pt}}|
    @{\hspace{1pt}}c
    @{\hspace{6pt}}c@{\hspace{6pt}}
    c@{\hspace{1pt}}|
    @{\hspace{1pt}}c
    @{\hspace{1pt}}c@{\hspace{1pt}}
    c|
    c
    @{\hspace{6pt}}c@{\hspace{6pt}}
    c@{\hspace{1pt}}|
    @{\hspace{1pt}}c
    @{\hspace{1pt}}c@{\hspace{1pt}}
    c|
    c
    @{\hspace{6pt}}c@{\hspace{6pt}}
    c@{\hspace{1pt}}|
    @{\hspace{1pt}}c
    @{\hspace{1pt}}c@{\hspace{1pt}}
    c@{\hspace{1pt}}
    }
      \hline
      \multirow{2}{*}{Criteria}                & \multirow{2}{*}{Config.}           & \multicolumn{6}{c|}{ResNet}                                                                                                              & \multicolumn{6}{c|}{VGG}                                                                                                              & \multicolumn{6}{c}{MobileNet}                                                                                                            \\
      \cline{3-20}
                                               &                                    & \texttt{CR} & $\texttt{CR}_{\times 10}$ & $\texttt{CR}_{\times 1}$ & \texttt{IN} & $\texttt{IN}_{\times 10}$ & $\texttt{IN}_{\times 1}$ & \texttt{CR} & $\texttt{CR}_{\times 10}$ & $\texttt{CR}_{\times 1}$ & \texttt{IN} & $\texttt{IN}_{\times 10}$ & $\texttt{IN}_{\times 1}$ & \texttt{CR} & $\texttt{CR}_{\times 10}$ & $\texttt{CR}_{\times 1}$ & \texttt{IN} & $\texttt{IN}_{\times 10}$ & $\texttt{IN}_{\times 1}$ \\
      \hline
      \multirow{3}{*}{\shortstack{NC ($\%$)}}  & \multirow{1}{*}{$T$=$0.25$}    & 0.026    & 0.031  & 0.031                                        & 0    & 0   & 0                         & 0.113   & 0.225  & 0.177                                        & 0   & 0   & 0           & 0         & 0       & 0                        & 0   & 0   & 0     \\
                            & \multirow{1}{*}{$T$=$0.5$}     & 0.75   & 2.10  & 1.14                                        & 0   & 0.01   & 0         & 0.56  & 1.09   & 0.77                                        & 0   & 0.01   & 0              & 0.02  & 0.05    & 0.01                                       & 0   & 0   & 0                     \\
                            & \multirow{1}{*}{$T$=$0.75$}     & 2.15   & 4.54  & 2.52                                        & 0.13   & 0.38   & 0.21     & 2.49  & 4.27   & 3.25                                        & 0.02   & 0.02   & 0       & 1.57   & 2.17  & 1.69                                        & 0.02   & 0.03   & 0.02                              \\
      \hline
      \multirow{3}{*}{\shortstack{KMNC ($\%$)}} & \multirow{1}{*}{$K$=$10$}    & 8.89  & 40.50    & 40.01                                       & 4.10   & 5.59   & 4.25      & 7.33  & 55.50   & 55.25                                       & 4.19   & 9.17   & 8.34          & 8.53  & 10.43   & 8.97                                        & 4.67   & 6.82   & 4.52                           \\
                            & \multirow{1}{*}{$K$=$100$}     & 8.52   & 28.37  & 26.96                                        & 2.17   & 5.14   & 5.08      & 7.54   & 45.60  & 45.25                                        & 1.36   & 5.26   & 5.22           & 4.81  & 10.19   & 10.02                                        & 1.43   & 3.01   & 2.15                          \\
                            & \multirow{1}{*}{$K$=$1,000$}     & 3.99  & 10.78   & 10.64                                       & 1.87   & 4.31   & 4.29       & 2.84   & 12.50  & 12.36                                       & 0.91   & 2.68   & 2.61       & 4.61   & 8.69  & 8.38                                         & 2.01   & 3.92    & 3.44                             \\
      \hline
      \multirow{1}{*}{\shortstack{NBC ($\%$)}}  & \multirow{1}{*}{N/A}     & 0    & 4.85     & 1.75                                        & 0.13  & 1.20   & 0.48     & 0.27   & 2.12   & 2.04                                        & 0.08  & 0.23   & 0.13         & 0.003   & 2.51  & 2.46                                        & 0.15  & 0.32   & 0.20                            \\
      \hline
      \multirow{1}{*}{\shortstack{SNAC ($\%$)}} & \multirow{1}{*}{N/A}    & 0    & 2.62     & 0.97                                        & 0.14  & 0.73   & 0.57         & 0.53   & 1.64   & 1.52                                        & 0.13  & 0.23   & 0.01                 & 0      & 2.37   & 1.41                                        & 0.15   & 0.38  & 0.27                \\
      \hline
      \multirow{3}{*}{\shortstack{TKNC ($\%$)}} & \multirow{1}{*}{$K$=$1$}     & 0.20   & 0.41  & 0.23                                        & 0.06  & 0.10   & 0.03         & 0.81   & 1.97  & 1.20                                        & 0.29  & 0.44   & 0.13      & 0.24   & 0.28  & 0.13                                        & 0.10  & 0.20   & 0.06                           \\
                            & \multirow{1}{*}{$K$=$10$}     & 0.38   & 0.68  & 0.36                                        & 0.11   & 0.14  & 0.02      & 2.29  & 3.83   & 2.66                                        & 0.01  & 0.01   & 0.01        & 0.21  & 0.31   & 0.12                                        & 0.03   & 0.03  & 0.03                            \\
                            & \multirow{1}{*}{$K$=$50$}     & 0.24  & 0.45   & 0.23                                        & 0.03  & 0.06  & 0.01    & 3.07   & 4.57  & 3.35                                        & 0.01  & 0.02   & 0.02         & 0.17  & 0.38   & 0.22                                        & 0.02  & 0.05   & 0                             \\
      \hline
      \multirow{3}{*}{\shortstack{TKNP ($\#$)}} & \multirow{1}{*}{$K$=$1$}     & 9,995  & 80,561   & 9,586                                       & 4,900  & 23,384  & 3,954      & 6,060  & 36,887   & 6,809                                        & 4,882  & 17,187  & 3,386           & 9,998   & 90,537  & 9,924                                       & 4,900  & 48,083  & 5,000                        \\
                            & \multirow{1}{*}{$K$=$10$}     & 10,000  & 100,000   & 10,000                                        & 4,901   & 50,000   & 5,000     & 10,000  & 100,000   & 10,000                                        & 4,901   & 50,000   & 5,000     & 10,000  & 100,000   & 10,000                                        & 4,901   & 50,000   & 5,000                           \\
                            & \multirow{1}{*}{$K$=$50$}     & 1,034  & 1,823   & 434                                        & 4,901   & 50,000   & 5,000   & 10,000  & 100,000  & 10,000                                   & 4,901   & 50,000   & 5,000           & 694    & 1,665   & 650                                         & 4,901   & 50,000   & 5,000                             \\
      \hline
      \multirow{3}{*}{\shortstack{CC* ($\#$)}}   & \multirow{1}{*}{\shortstack{$T$=$10$ / $T$=$1,000$}} & 55  & 60   & 44                                      & 4,945   & 9,787   & 1,333           & 147  & 168  & 137                                     & \colorbox{pptgreen3!25}{14,532}  & \colorbox{pptgreen3!25}{7,681}   & \colorbox{pptgreen3!25}{2,148}        & 37   & 38  & 29                                       & 4,947   & 43,613   & 4,614                       \\
                            & \multirow{1}{*}{\shortstack{$T$=$20$ / $T$=$2,000$}}   & 19   & 23  & 19                                       & 4,918  & 9,771  & 397         & 51  & 63   & 48                                       & \colorbox{pptgreen3!25}{11,914 } & \colorbox{pptgreen3!25}{1,606} & \colorbox{pptgreen3!25}{954}    & 9   & 11   & 10                                       & 4,946  & 12,466  & 2,618                              \\
                            & \multirow{1}{*}{\shortstack{$T$=$50$ / $T$=$5,000$}}     & 9  & 9   & 7                                        & 4,130   & 8,263   & 256          & 22  & 22  & 21                                      & \colorbox{pptgreen3!25}{4,526}   & \colorbox{pptgreen3!25}{571}   & \colorbox{pptgreen3!25}{426}   & 7   & 7  & 4                                        & \colorbox{pptgreen3!25}{4,943}   & \colorbox{pptgreen3!25}{531}  & \colorbox{pptgreen3!25}{346}                              \\
      \hline
      \multirow{3}{*}{\shortstack{LSC** ($\#$)}}   & \multirow{1}{*}{\shortstack{$T$=$1$ / $T$=$100$}}  & 196   & 628  & 423                                      & 1,475   & 2095   & 1,373       & 400  & 1,153  & 671                                     & 1,243  & 2,070   & 1,028       & 164  & 553   & 574                                       & 2,028   & 3,118   & 1,858                             \\
                            & \multirow{1}{*}{\shortstack{$T$=$10$ / $T$=$500$}}   & 44   & 69  & 83                                       & 414  & 502  & 420        & 79   & 117   & 98                                       & 357  & 447  & 275        & 36   & 51   & 48                                       & 610  & 755  & 545                           \\
                            & \multirow{1}{*}{\shortstack{$T$=$100$ / $T$=$1,000$}}     & 7  & 7   & 8                                        & 227   & 214   & 228    & 10  & 13  & 12                                      & 199   & 223   & 201    & 5  & 7   & 6                                        & 256   & 359   & 335                                   \\
      \hline
      \multirow{3}{*}{\shortstack{DSC ($\#$)}} & \multirow{1}{*}{$T$=$0.01$}     & 489  & 3,478   & 831                                       & \multicolumn{3}{c|}{Failed}   & 566  & 6,431   & 1,120                                        & \multicolumn{3}{c|}{Failed}       & 471   & 4,491   & 1,742                                      & \multicolumn{3}{c}{Failed}            \\
                            & \multirow{1}{*}{$T$=$0.1$}     & 146    & 426  & 307                                       & \multicolumn{3}{c|}{Failed}   & 253   & 4,983  & 565                                       & \multicolumn{3}{c|}{Failed}   & 148  & 1,276   & 460                                        & \multicolumn{3}{c}{Failed}                \\
                            & \multirow{1}{*}{$T$=$1$}     & 34  & 39   & 54                                        & \multicolumn{3}{c|}{Failed}   & 107  & 1,195  & 110                                   & \multicolumn{3}{c|}{Failed}     & 42   & 161     & 75                                        & \multicolumn{3}{c}{Failed}              \\
      \hline
      \multirow{3}{*}{\shortstack{MDSC ($\#$)}} & \multirow{1}{*}{$T$=$1$}     & 422  & 987    & 675                                      & \multicolumn{3}{c|}{Failed}    & 639   & 4,065  & 1,538                                        & \multicolumn{3}{c|}{Failed}      & 651  & 1,023    & 733                                      & \multicolumn{3}{c}{Failed}            \\
                            & \multirow{1}{*}{$T$=$10$}     & 73  & 93    & 100                                       & \multicolumn{3}{c|}{Failed}    & 208  & 780   & 365                                        & \multicolumn{3}{c|}{Failed}       & 176  & 400   & 279                                        & \multicolumn{3}{c}{Failed}           \\
                            & \multirow{1}{*}{$T$=$100$}     & 7  & 10   & 9                                        & \multicolumn{3}{c|}{Failed}    & 50  & 110    & 87                                  & \multicolumn{3}{c|}{Failed}       & 38    & 31   & 42                                         & \multicolumn{3}{c}{Failed}           \\
      \hline
      \multirow{1}{*}{\textbf{NLC}} & \multirow{1}{*}{N/A}     & \colorbox{pptgreen3!25}{1.03}  & \colorbox{pptgreen3!25}{0.47}  & \colorbox{pptgreen3!25}{0.05}                                  & \colorbox{pptgreen3!25}{6006.38}   & \colorbox{pptgreen3!25}{113.37}   & \colorbox{pptgreen3!25}{11.00}      & \colorbox{pptgreen3!25}{403.95} & \colorbox{pptgreen3!25}{0.004} & \colorbox{pptgreen3!25}{0}                                     & \colorbox{pptgreen3!25}{20327.50}   & \colorbox{pptgreen3!25}{413.50}   & \colorbox{pptgreen3!25}{49.50}     & \colorbox{pptgreen3!25}{0.61}  & \colorbox{pptgreen3!25}{0.23}  & \colorbox{pptgreen3!25}{0.12}                                  & \colorbox{pptgreen3!25}{10693.50}   & \colorbox{pptgreen3!25}{757.25}   & \colorbox{pptgreen3!25}{86.25}                                \\
      \hline
  \end{tabular}
  }
  \begin{tablenotes}
      \footnotesize
      \item *$T \in \{10, 20, 50\}$ for CIFAR10 (\texttt{CR}) and $T \in \{1000, 2000, 5000\}$ for ImageNet (\texttt{IN}).
      \item **$T \in \{1, 10, 100\}$ for CIFAR10 (\texttt{CR}) and $T \in \{100, 500, 1000\}$ for ImageNet (\texttt{IN}).
  \end{tablenotes}
  \vspace{-15pt}
\end{table*}

\begin{table}[t]
  \caption{Coverage achieved by different datasets on generative models.
    $\texttt{CR}$ and $\texttt{IN}$ refer to BigGAN
    trained on CIFAR10 and ImageNet. Assessments matching ground truth are
    \colorbox{pptgreen3!25}{marked}.}
  \label{tab:gen-rq1}
  \centering
\resizebox{0.8\linewidth}{!}{
  \begin{tabular}{
    @{\hspace{1pt}}c@{\hspace{1pt}}|
    @{\hspace{1pt}}c@{\hspace{1pt}}|
    @{\hspace{1pt}}c@{\hspace{1pt}}|
    @{\hspace{1pt}}c@{\hspace{1pt}}|
    @{\hspace{1pt}}c@{\hspace{1pt}}|
    @{\hspace{1pt}}c@{\hspace{1pt}}|
    @{\hspace{1pt}}c@{\hspace{1pt}}|
    @{\hspace{1pt}}c@{\hspace{1pt}}
    }
      \hline
      Criteria              & Config.              & DNN                             & Normal & ${N/2}$ & ${N/10}$ & ${C/2}$ & ${C/10}$ \\
      \hline
      \multirow{6}{*}{\shortstack{NC\\($\%$)}}  & \multirow{2}{*}{$T$=$0.25$} & $\texttt{CR}$ & 99.97    & 99.97               & 99.95               & 99.97 & 99.95                          \\
                            &                             & $\texttt{IN}$ & 99.43    & 99.42               & 99.40               & 99.42 & 99.39                          \\
                            & \multirow{2}{*}{$T$=$0.5$} & $\texttt{CR}$ & 99.34    & 99.30               & 99.02               & 99.25 & 98.73                          \\
                            &                            & $\texttt{IN}$ & 94.18    & 93.96               & 93.44               & 93.87 & 93.44                          \\
                            & \multirow{2}{*}{$T$=$0.75$} & $\texttt{CR}$ & 86.66    & 84.88               & 79.70               & 82.85 & 76.90                          \\
                            &                             & $\texttt{IN}$ & 81.64    & 81.30               & 80.17               & 80.98 & 77.61                          \\
      \hline
      \multirow{6}{*}{\shortstack{KMNC\\($\%$)}} & \multirow{2}{*}{$K$=$10$} & $\texttt{CR}$ & 54.72    & 56.00               & 54.71               & 54.24 & 52.44                           \\
                            &                           & $\texttt{IN}$ & 58.15    & 57.06               & 58.14               & 56.20 & 56.82                          \\
                            & \multirow{2}{*}{$K$=$100$} & $\texttt{CR}$ & 21.54    & 21.45               & 21.84               & 21.83 & 21.48                          \\
                            &                            & $\texttt{IN}$ & 22.66    & 22.33               & 22.59               & 22.20 & 22.32                          \\
                            & \multirow{2}{*}{$K$=$1000$} & $\texttt{CR}$ & 3.04    & 3.04               & 3.04               & 3.04 & 3.03                          \\
                            &                             & $\texttt{IN}$ & 3.07    & 3.08               & 3.07               & 3.07 & 3.06                          \\
      \hline
      \multirow{2}{*}{\shortstack{NBC\\($\%$)}}  & \multirow{2}{*}{N/A} & $\texttt{CR}$ & 0.22    & 0.09               & 0.04               & 0.09 & 0.10                          \\
                            &                      & $\texttt{IN}$ & 0.54    & 0.21               & 0.21               & 0.03 & 0.06                          \\                    
      \hline
      \multirow{2}{*}{\shortstack{SNAC\\($\%$)}} & \multirow{2}{*}{N/A} & $\texttt{CR}$ & 0.02    & 0.03               & 0.03               & 0.03 & 0.03                          \\
                            &                      & $\texttt{IN}$ & 0.03    & 0.07               & 0.04               & 0.64 & 0.14                          \\
      \hline
      \multirow{6}{*}{\shortstack{TKNC\\($\%$)}} & \multirow{2}{*}{$K$=$1$} & $\texttt{CR}$ & 31.69    & 27.75               & 16.95               & 27.16 & 15.56                          \\
                            &                          & $\texttt{IN}$ & 16.79    & 14.56               & 9.62               & 13.11 & 5.91                          \\
                            & \multirow{2}{*}{$K$=$10$} & $\texttt{CR}$ & 62.64    & 58.89               & 48.34               & 58.46 & 45.14                          \\
                            &                           & $\texttt{IN}$ & 34.92    & 33.77               & 29.90               & 32.44 & 22.62                          \\
                            & \multirow{2}{*}{$K$=$50$} & $\texttt{CR}$ & 83.96    & 81.45               & 75.03               & 81.23 & 73.54                          \\
                            &                           & $\texttt{IN}$ & 42.02    & 41.68               & 40.58               & 41.27 & 38.08                          \\
      \hline
      \multirow{6}{*}{\shortstack{TKNP\\($\#$)}} & \multirow{2}{*}{$K$=$1$} & $\texttt{CR}$ & 31,415    & 15,843               & 3,193               & 15,779 & 3,134                          \\
                            &                          & $\texttt{IN}$ & 32,000    & 16,000               & 3,200               & 16,000 & 3,200                          \\
                            & \multirow{2}{*}{$K$=$10$} & $\texttt{CR}$ & 32,000    & 16,000               & 3,200               & 16,000 & 3,200                          \\
                            &                           & $\texttt{IN}$ & 32,000    & 16,000               & 3,200               & 16,000 & 3,200                          \\
                            & \multirow{2}{*}{$K$=$50$} & $\texttt{CR}$ & 32,000    & 16,000               & 3,200               & 16,000 & 3,200                          \\
                            &                           & $\texttt{IN}$ & 10,894    & 6,898               & 2,088               & 6,997 & 2,178                          \\
      \hline
      \multirow{6}{*}{\shortstack{CC\\($\#$)}}  & \multirow{2}{*}{$T$=$50$} & $\texttt{CR}$ & 347    & 302               & 175               & 232  & 89                         \\
                            &                          & $\texttt{IN}$ & 1,204    & 921               & 485               & 780 & 324                          \\
                            & \multirow{2}{*}{$T$=$100$} & $\texttt{CR}$ & 90    & 70               & 40               & 50 & 18                          \\
                            &                            & $\texttt{IN}$ & 323    & 266               & 147               & 239 & 108                          \\
                            & \multirow{2}{*}{$T$=$200$} & $\texttt{CR}$ & 27    & 25               & 13               & 23 & 7                          \\
                            &                            & $\texttt{IN}$ & 105    & 99               & 77               & 95 & 64                          \\
      \hline
      \multirow{6}{*}{\shortstack{LSC\\($\#$)}}  & \multirow{2}{*}{\shortstack{$T$=$1$\\$T$=$100$}} & $\texttt{CR}$ & 1,286    & 1,047               & 627               & 1,011  & 607                         \\
                            &                          & $\texttt{IN}$ & 197    & 154               & 63               & 153 & 48                          \\
                            & \multirow{2}{*}{\shortstack{$T$=$10$\\$T$=$500$}} & $\texttt{CR}$ & 216    & 188               & 136               & 181 & 122                          \\
                            &                            & $\texttt{IN}$ & 92    & 68               & 27               & 61 & 27                          \\
                            & \multirow{2}{*}{\shortstack{$T$=$100$\\$T$=$1,000$}} & $\texttt{CR}$ & 40    & 32               & 20               & 37 & 21                          \\
                            &                            & $\texttt{IN}$ & 66    & 42               & 14               & 43 & 17                          \\
      \hline
      \multirow{6}{*}{\shortstack{DSC\\($\#$)}}  & \multirow{2}{*}{$T$=$0.01$} & $\texttt{CR}$ & 123    & 111               & 90               & 90  & 74                         \\
                            &                          & $\texttt{IN}$ & 412    & 325               & 184               & 317 & 167                          \\
                            & \multirow{2}{*}{$T$=$0.1$} & $\texttt{CR}$ & 13    & 13               & 12               & 12 & 10                          \\
                            &                            & $\texttt{IN}$ & 76    & 56               & 39               & 67 & 31                          \\
                            & \multirow{2}{*}{$T$=$1$} & $\texttt{CR}$ & 2    & 2               & 2               & 2 & 2                          \\
                            &                            & $\texttt{IN}$ & 10    & 9               & 7               & 10 & 7                          \\
      \hline
      \multirow{6}{*}{\shortstack{MDSC\\($\#$)}}  & \multirow{2}{*}{$T$=$1$} & $\texttt{CR}$ & 249    & 232               & 97               & 260  & 89                         \\
                            &                          & $\texttt{IN}$ & 943    & 673               & 418               & 640 & 401                          \\
                            & \multirow{2}{*}{$T$=$10$} & $\texttt{CR}$ & 24    & 19               & 18               & 17 & 24                          \\
                            &                            & $\texttt{IN}$ & 224    & 188               & 119               & 222 & 122                          \\
                            & \multirow{2}{*}{$T$=$100$} & $\texttt{CR}$ & 3    & 4               & 3               & 2 & 2                          \\
                            &                            & $\texttt{IN}$ & 119    & 47               & 15               & 45 & 32                          \\
      \hline
\multirow{2}{*}{\textbf{\tool}}& \multirow{2}{*}{N/A} & $\texttt{CR}$  & \colorbox{pptgreen3!25}{1051.63}    & \colorbox{pptgreen3!25}{776.29}               & \colorbox{pptgreen3!25}{653.62}               & \colorbox{pptgreen3!25}{280.67} & \colorbox{pptgreen3!25}{178.92}                          \\
                            &                      & $\texttt{IN}$ & \colorbox{pptgreen3!25}{3596.89}    & \colorbox{pptgreen3!25}{3306.88}               & \colorbox{pptgreen3!25}{3141.02}               & \colorbox{pptgreen3!25}{2636.64} & \colorbox{pptgreen3!25}{1462.11}                          \\
      \hline
  \end{tabular}
}
\vspace{-10pt}
\end{table}

\noindent \textbf{Generative Models.}~Generative models generate images
by sampling from a fixed distribution (i.e., the normal distribution). Since this procedure is
conditioned on a given class label, we construct datasets of different diversity
by following the $Normal$, $N/x$ and $C/x$ schemes (see \T~\ref{tab:gen-rq1}).
For the $Normal$ scheme, we sample total $m$ inputs from the normal distribution
and each class has the same number of inputs. For the $N/x$ scheme, we only keep
$1/x$ of inputs from each class, whereas for the $C/x$ scheme, we only keep
$1/x$ of classes. We set $m$ to $32,000$ and choose $x \in \{2, 10\}$.

\noindent \underline{Ground Truth (\T~\ref{tab:gen-rq1}):}~Since inputs conditioned on different classes
are all sampled from normal distribution, behaviors of generative models are
primarily decided by the class labels. Thus, the diversity of these datasets
are ranked as: \underline{$Normal > N/2 > N/10 > C/2 > C/10$}.

\noindent \textbf{Result Overview.}~We show the increased coverage of different
test suites compared with training data (since the DNN is already ``tested'' using
its training data) in \T~\ref{tab:text-rq1}, \T~\ref{tab:disc-rq1} and \T~\ref{tab:gen-rq1}
for text (discriminative) models, image (discriminative) models, and generative models,
respectively. All experiments are launched for five times to reduce randomness.
The maximum deviations are around $0.01\%$ and $1$ for criteria in $\%$ and $\#$ units,
respectively.

\noindent \textbf{Scalability Issues of MDSC/DSC.}~In \T~\ref{tab:disc-rq1}, we only evaluate MDSC/DSC using CIFAR10. MDSC
requires storing class-conditional covariance matrices, whose space cost is
$O(\text{\#class} \times {\text{\#neuron}}^2)$. Note that ImageNet has 1K
classes and 1M training data (\T~\ref{tab:data-rq1}).
DSC takes over 12 hours to assess one scheme of ImageNet (criteria satisfying
\CircleSeven\ only take several minutes) since it iterates over 1M neuron output
traces for each test input. It's thus infeasible to apply MDSC/DSC on models
trained on ImageNet that is much larger than CIFAR10.

\noindent \textbf{Impacts of Hyper-parameters.}~Previous criteria (except for NBC
and SNAC) require extra hyper-parameters, which largely affect the results. For
instance, in \T~\ref{tab:disc-rq1}, all variants of CIFAR10 test data have \textit{nearly zero} coverage when $T$
of NC is $0.25$, whereas they reach a saturation coverage when $K$ of TKNP is
$10$ (i.e., \#patterns equals to \#inputs) despite they are still evaluated on
same models. Moreover, the hyper-parameter $T$ of CC is model specific, for
instance, $T=10$ is suitable for ResNet trained on CIFAR10, but is too small if
the ResNet is trained on ImageNet --- any input will form a new cluster. We thus
increase $T$ 100 times for models trained on ImageNet. SCs suffer from the
similar issue (e.g., LSC in \T~\ref{tab:disc-rq1}) even though we have optimized
the hyper-parameter configurations. Before applying these criteria, users need to
scope the distance range for each model. Again, without prior knowledge on setting
hyper-parameters, it is obscure to adopt these criteria in real-life usage.
In contrast, \tool\ is \textit{hyper-parameter free}, enabling an ``out-of-the-box'' usage.

\noindent \textbf{Discriminative Model: Quantity, not Diversity.}~For most cases in \T~\ref{tab:text-rq1} and \T~\ref{tab:disc-rq1},
the $\times 1$ scheme and the test dataset have comparable results, and the $\times 10$
scheme has the highest coverage. Given datasets created by $\times 1$ have an identical size with the test
dataset, we interpret that previous neuron coverage criteria are more sensitive
to the \textit{number} of data samples rather than the diversity. Nevertheless,
the coverage assessed by \tool\ is consistent with the diversity of these
datasets (i.e., the ground truth). Moreover, despite the number of $\times 10$
variant is 10 times of test data, its reflected coverage on \tool\ is smaller,
indicating that \tool\ successfully reflects input diversity.

\noindent \textbf{Discriminative Model: Dissimilarity, not Diversity.}~We note that in most
cases, SCs have $\text{test}_{\times 1} > \text{test}$. Recall that SCs primarily
describe how distinct a test input is towards training data. This implies that SCs
are more sensitive to the dissimilarity rather than the diversity of test suites.
Also, though CC and SCs are conceptually similar to some extent, we find that CC
has relatively better performance. CC represents coverage as \#clusters based on
Euclidean distance, in contrast, SCs focus on designing new distance metrics.
To compute coverage, SCs simply discretize the distance values as buckets.
From this comparison, it might not be inaccurate to interpret that how to represent
coverage based on the distance may be more essential than the distance metric itself.

\noindent \textbf{Generative Models: the Incapability.}~To train generative
models like BigGAN, inputs for both training and test phases are random variables
sampled from the normal distribution (here, training images act as ``labels'').
Infinite ``training inputs'' exist, and therefore, KMNC, NBC, SNAC and SCs are
infeasible to use, as they require to first obtain neuron outputs using all training
inputs. Practically, we use all samples from $Normal$ schemes to simulate the
training inputs. Nevertheless, we find that coverage values measured by them are
random (e.g., the ``\texttt{CR}'' row of SNAC in \T~\ref{tab:gen-rq1}). In contrast, \tool,
by default, can be used with \textit{no} training data, and it can be adopted in
both discriminative and generative models.

\noindent \textbf{Generative Models: Quantity, not Diversity.}~For the remaining coverage criteria, most of their results
are ranked as $Normal > N/2 > C/2 > N/10 > C/10$, which further reflects that
prior criteria are more sensitive to the number of inputs instead of their
diversity. Since previous criteria consider neurons separately, feeding more
inputs might increase the chance of activating more neurons or producing a
distinct trace of neuron outputs. In contrast, \tool\ analyzes the
entanglements among neurons, where most ``trivial'' inputs will not influence
the correlations or distributions. Since a diverse test suite possibly uncovers
different neuron correlations, \tool\ becomes more responsible for a diverse
test suite.

\begin{figure}[!ht]
  \centering
  \hspace{-15pt}
  \includegraphics[width=1.05\linewidth]{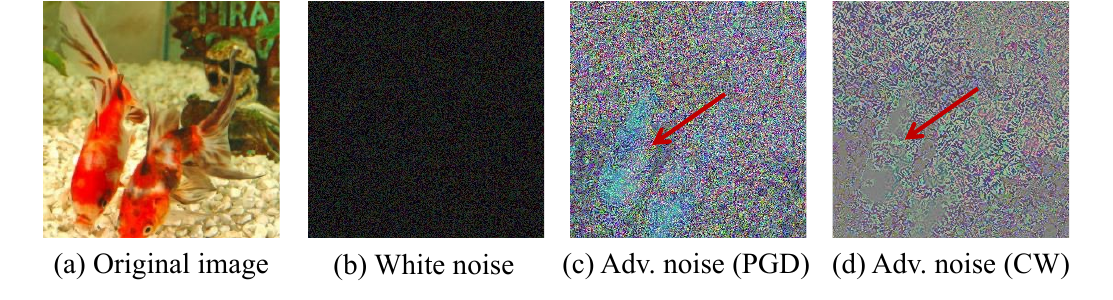}
  \vspace{-20pt}
  \caption{Adversarial perturbations.}
  \vspace{-10pt}
  \label{fig:adv}
\end{figure}

\begin{table}[t]
  \caption{Coverage achieved by test data and AEs of CIFAR10$\dagger$. Assessments
  matching ground truth are \colorbox{pptgreen3!25}{marked}.}
   \vspace{-5pt}
  \label{tab:fault-rq1}
  \centering
\resizebox{1.0\linewidth}{!}{
  \begin{tabular}{
    @{\hspace{1pt}}c@{\hspace{1pt}}|
    @{\hspace{1pt}}c@{\hspace{1pt}}|
    @{\hspace{1pt}}c@{\hspace{1pt}}|
    @{\hspace{1pt}}c@{\hspace{1pt}}|
    @{\hspace{1pt}}c@{\hspace{1pt}}|
    @{\hspace{1pt}}c@{\hspace{1pt}}|
    @{\hspace{1pt}}c@{\hspace{1pt}}|
    @{\hspace{1pt}}c@{\hspace{1pt}}|
    @{\hspace{1pt}}c@{\hspace{1pt}}|
    @{\hspace{1pt}}c@{\hspace{1pt}}|
    @{\hspace{1pt}}c@{\hspace{1pt}}|
    @{\hspace{1pt}}c@{\hspace{1pt}}|
    @{\hspace{1pt}}c@{\hspace{1pt}}
    }
    \hline
    \multirow{3}{*}{}       & \multirow{3}{*}{Data}       & NC        & KMNC      & \multirow{2}{*}{NBC} & \multirow{2}{*}{SNAC} & TKNC     & TKNP     & CC       & LSC      & DSC      & MDSC     & \multirow{3}{*}{\textbf{NLC}} \\
                            &                        & $0.75$ & $100$ &                      &                       & $10$ & $50$ & $10$ & $10$ & $0.1$ & $10$ & \\
                            &                        & ($\%$)     & ($\%$)    &     ($\%$)           &        ($\%$)         & ($\%$)   & ($\#$)   & ($\#$)   & ($\#$)   & ($\#$)   & ($\#$)   & \\
    \hline
    \multirow{3}{*}{R} & Test                     & 2.15      & 8.52      & 0                    & 0                     & 0.38   & 1,034    & 55  & 44 & 146 & 73 & 1.0 \\
                            & CW                       & 0.90      & 0         & \colorbox{pptgreen3!25}{3.94}                 & \colorbox{pptgreen3!25}{2.73}             & 0.18   & 31       & 5   & 0 & \colorbox{pptgreen3!25}{336} & 0 & \colorbox{pptgreen3!25}{2.4} \\
                            & PGD                      & 1.43      & 0         & \colorbox{pptgreen3!25}{13.83}                & \colorbox{pptgreen3!25}{23.22}            & 0.35   & 680      & 29  & \colorbox{pptgreen3!25}{115} & \colorbox{pptgreen3!25}{785} & \colorbox{pptgreen3!25}{372} & \colorbox{pptgreen3!25}{22.9} \\
    \hline
    \multirow{3}{*}{V}    & Test                     & 2.49      & 7.54      & 0.27                 & 0.53                  & 2.29   & 10,000   & 147 & 79 & 253 & 208 & 403.9 \\
                            & CW                       & 0.20      & 0         & \colorbox{pptgreen3!25}{5.03}                 & \colorbox{pptgreen3!25}{7.08}                & 0.20   & 9,542    & 125   & 0 & \colorbox{pptgreen3!25}{949} & 0 & \colorbox{pptgreen3!25}{931.0} \\
                            & PGD                      & 1.69      & 0.74      & \colorbox{pptgreen3!25}{12.64}                & \colorbox{pptgreen3!25}{15.23}               & 1.32   & 4,997    & 142  & \colorbox{pptgreen3!25}{222} & \colorbox{pptgreen3!25}{5,901} & \colorbox{pptgreen3!25}{1,461} & \colorbox{pptgreen3!25}{4749.1} \\
    \hline
    \multirow{3}{*}{M} & Test                  & 1.57      & 4.81      & 0                    & 0                     & 0.21   & 694      & 37  & 36 & 148 & 176 & 0.6 \\
                               & CW                    & 0.02      & 0         & \colorbox{pptgreen3!25}{1.77}                 & \colorbox{pptgreen3!25}{1.62}          & 0.14   & 172      & 13   & 0 & \colorbox{pptgreen3!25}{452} & 0 & \colorbox{pptgreen3!25}{0.9} \\
                               & PGD                   & 0.36      & 0         & \colorbox{pptgreen3!25}{9.71}                 & \colorbox{pptgreen3!25}{10.39}         & 0.16   & 299      & 16   & \colorbox{pptgreen3!25}{65} & \colorbox{pptgreen3!25}{3,648} & 97 & \colorbox{pptgreen3!25}{13.6} \\
    \hline
  
  \end{tabular}
  }
  \begin{tablenotes}
    \footnotesize
    \item $\dagger$ R, V, M denote ResNet, VGG, MobileNet, respectively.
  \end{tablenotes}
  \vspace{-5pt}
\end{table}

\begin{table}[t]
  \caption{Coverage achieved by AEs generated (via PGD) using test data of CIFAR10$\dagger$.
  Assessments matching ground truth are \colorbox{pptgreen3!25}{marked}.}
   \vspace{-5pt}
  \label{tab:fault-test-rq1}
  \centering
\resizebox{1.0\linewidth}{!}{
  \begin{tabular}{
    @{\hspace{1pt}}c@{\hspace{1pt}}|
    @{\hspace{1pt}}c@{\hspace{1pt}}|
    @{\hspace{1pt}}c@{\hspace{1pt}}|
    @{\hspace{1pt}}c@{\hspace{1pt}}|
    @{\hspace{1pt}}c@{\hspace{1pt}}|
    @{\hspace{1pt}}c@{\hspace{1pt}}|
    @{\hspace{1pt}}c@{\hspace{1pt}}|
    @{\hspace{1pt}}c@{\hspace{1pt}}|
    @{\hspace{1pt}}c@{\hspace{1pt}}|
    @{\hspace{1pt}}c@{\hspace{1pt}}|
    @{\hspace{1pt}}c@{\hspace{1pt}}|
    @{\hspace{1pt}}c@{\hspace{1pt}}
    }
    \hline
     \multirow{3}{*}{Model}       & NC        & KMNC      & \multirow{2}{*}{NBC} & \multirow{2}{*}{SNAC} & TKNC     & TKNP     & CC       & LSC      & DSC      & MDSC     & \multirow{3}{*}{\textbf{NLC}} \\
                            & $0.75$ & $100$ &                      &                       & $10$ & $50$ & $10$ & $10$ & $0.1$ & $10$ & \\
                            & ($\%$)     & ($\%$)    &     ($\%$)           &        ($\%$)         & ($\%$)   & ($\#$)   & ($\#$)   & ($\#$)   & ($\#$)   & ($\#$)   & \\
    \hline
     ResNet                    & 0.41      & 0      & \colorbox{pptgreen3!25}{9.17}  & \colorbox{pptgreen3!25}{13.04}           & 0.16   & 980    & 20  & 0 & \colorbox{pptgreen3!25}{590}   & \colorbox{pptgreen3!25}{323} & \colorbox{pptgreen3!25}{8.60} \\
     VGG                       & 0.58      & 0      & \colorbox{pptgreen3!25}{12.40} & \colorbox{pptgreen3!25}{14.90}           & 0.44   & 4,999  & 94  & 0 & \colorbox{pptgreen3!25}{3,044} & \colorbox{pptgreen3!25}{917} & \colorbox{pptgreen3!25}{1584.4} \\
     MobileNet                 & 0.20      & 0      & \colorbox{pptgreen3!25}{2.30}  & \colorbox{pptgreen3!25}{3.10}            & 0.17   & 639    & 12  & 0 & \colorbox{pptgreen3!25}{2,808} & 86                           & \colorbox{pptgreen3!25}{10.43} \\
    \hline
  
  \end{tabular}
  }
  \vspace{-10pt}
\end{table}

\subsubsection{Fault-Revealing Capability of Test Suites}

\noindent \textbf{Setup.}~To evaluate the fault-revealing
capabilities of test suites, we construct adversarial examples (AE) using two
adversarial attack algorithms, Carlini/Wagner (CW)~\cite{carlini2017cw} and
Project Gradient Descent (PGD)~\cite{madry2017pgd}. CW and PGD are two
representative types of algorithms for AE generation, namely, optimization-based
and loss function-based.
Perturbations added on AEs are different from white noise, as
shown in \F~\ref{fig:adv}, which generally highlight objects in images.
AEs are generated using the training data (see ``\textbf{Clarification}'' below
for results of AEs generated using test data),
and all algorithms attacking the three models reach over $98\%$ success rates.
That is, more than $0.98 \times \text{(\#training data)}$ AEs can trigger erroneous
behaviors. The incorrect predictions uniformly distribute across all classes.

\noindent \underline{Ground Truth (\T~\ref{tab:fault-rq1}):} Since the AE set manifests a higher
fault-revealing capability (also discloses more untested cases), it should
have a higher coverage value. Thus, the ground in this evaluation
is \underline{PGD $>$ Test}, or \underline{CW $>$ Test}.

\noindent \textbf{Result Overview.}~As with the previous settings, we record
and present the increased coverage of test data and AEs in
\T~\ref{tab:fault-rq1}. NC, TKNC, TKNP and CC have higher coverage for the test
data than AEs, indicating that they are unable to assess the fault-revealing
capability of a test suite. Harel-Canada et al.~\cite{harel2020neuron} also find
that NC is not strongly correlated with defect detection, which is aligned with
our results. Nevertheless, \T~\ref{tab:fault-rq1} shows that \tool\ is
strongly and positively correlated to the fault-revealing capability of a test
suite.

\noindent \textbf{Implementation Issues of LSC.}~MDSC has zero coverage
on AEs generated using CW. Generally, perturbations added by CW are harder to
identify~\cite{carlini2017cw}. This statement can also be illustrated by
\T~\ref{tab:fault-rq1} (nearly all coverage values of PGD are much higher than CW)
and \F~\ref{fig:adv}. Though LSC also gives zero coverage for CW AEs, it is
seemingly due to some numerical issues. In particular, we observed some
\texttt{inf} (infinite numbers) when LSC was computing the Gaussian kernel to
measure the ``surprise'' of a test input. AE may induce a large neuron output,
which consequently result in \texttt{inf} since exponential functions
($f(x) = e^x$) are used by the Gaussian kernel. DSC shows generally promising
results for this evaluation, giving higher coverage values for AEs than normal
test inputs.

\noindent \textbf{Correlation to ``Out-of-Bound'' Neuron Outputs.}~KMNC has
no response to AEs, but NBC and SNAC respond notably, because AEs
primarily lead to out-of-range neuron outputs. \tool\ measures how
divergent a layer output is and responds to the density change of layer outputs.
Therefore, the out-of-range neuron outputs, which either induce a
higher variance or introduce a low-density layer output, can be captured.

\noindent \textbf{Clarification.}~Using training data to generate AEs helps
alleviate side effects caused by the diversity of test data. To clarify, AEs
generated using training data should not introduce new semantical content,
because AEs retain patterns in the training data except several stealthily
mutated pixels (e.g., a cat in a seed image still appears in the resulting AE).
Therefore, if the coverage increases, it shows the fault-revealing capability.
Nevertheless, if we use AEs generated by the test dataset, it is hard to
determine if the increase is due to unseen semantical content on AEs (because
test inputs are unseen by DNNs), or due to the fault-revealing capability. 
We also report results on AEs generated using test data in
\T~\ref{tab:fault-test-rq1}, whose findings are generally consistent with
\T~\ref{tab:fault-rq1}.

\begin{table}[t]
  \caption{Coverage achieved by adversarial-perturbed inputs (AP; not AE)
  generated using test data of CIFAR10$\dagger$.
  Assessments matching ground truth are \colorbox{pptgreen3!25}{marked}.}
   \vspace{-5pt}
  \label{tab:fault-faithfulness-rq1}
  \centering
\resizebox{1.0\linewidth}{!}{
  \begin{tabular}{
    @{\hspace{1.5pt}}c@{\hspace{1.5pt}}|
    c
    c
    @{\hspace{1.5pt}}c@{\hspace{1.5pt}}
    c|
    c
    @{\hspace{4pt}}c@{\hspace{1.5pt}}
    c
    c
    }
    \hline
                       & \multicolumn{4}{c|}{PGD} & \multicolumn{4}{c}{CW} \\
    \cline{2-9}
    \multirow{3}{*}{}  & \multirow{2}{*}{NBC} & \multirow{2}{*}{SNAC} & DSC            & \multirow{3}{*}{\textbf{\tool}}           & \multirow{2}{*}{NBC} & \multirow{2}{*}{SNAC} & DSC                                  & \multirow{3}{*}{\textbf{\tool}} \\
                       &                      &                       & $0.1$          &                                         &                      &                       & $0.1$                                & \\
                       &     ($\%$)           &        ($\%$)         & ($\#$)         &                                         &      ($\%$)          &        ($\%$)         & ($\#$)                               & \\
    \hline
    ResNet             & 0                    & 0                     & 463 $>$ Test   & \colorbox{pptgreen3!25}{0.20}  & 3.33 $>$ Test        & 2.30 $>$ Test         & 0                                    & \colorbox{pptgreen3!25}{0.13} \\
    VGG                & 0                    & 0                     & 1,295 $>$ Test & \colorbox{pptgreen3!25}{48.35} & 2.06 $>$ Test        & 4.30 $>$ Test         & \colorbox{pptgreen3!25}{52} & \colorbox{pptgreen3!25}{36.21} \\
    MobileNet          & 0                    & 0                     & 427 $>$ Test   & \colorbox{pptgreen3!25}{0.06}  & 1.12 $>$ Test        & 1.70 $>$ Test         & 0                                    & \colorbox{pptgreen3!25}{0.02} \\
    \hline
  
  \end{tabular}
  }
  \vspace{-10pt}
\end{table}

\noindent \textbf{Adversarial Perturbations.}~While NBC/SNAC/DSC appear to match
all ground truth in \T~\ref{tab:fault-rq1}, we suspect that they respond to the
adversarial perturbations instead of the fault-revealing capability of AEs. This
is derived from the following controlled experiments. We first randomly pick
1,000 inputs from the test set (which has 10,000 inputs). Then, we add
adversarial-perturbations (CW/PGD) on these inputs but do \textit{not} change
their predictions (they are thus \textit{not} AE). Let these
adversarial-perturbed inputs be AP. We compare the increased coverage of AP vs.
test dataset (Test).

The \underline{ground truth (\T~\ref{tab:fault-faithfulness-rq1})} should be:
\underline{$0 < \text{AP} < \text{Test}$}. AP $<$ Test because AP do not lead to
mispredictions, but Test is ten times larger than AP. 0 $<$ AP because AP has
1,000 inputs of diverse semantical contents.
Results are in \T~\ref{tab:fault-faithfulness-rq1}. \tool\ has correct results for all
settings, and DSC only matches ground truth for $\langle \text{VGG} \& \text{CW} \rangle$.
NBC/SANC fail in all settings.

This is an important observation, indicating that NBC/SNAC/DSC do not accurately
reflect the fault-revealing capability. Instead, they are sensitive to the
``dissimilarity'' induced by adversarial perturbations (even if Test is
10$\times$ larger than AP). As introduced in \S~\ref{subsec:nc}, NBC/SNAC
increase if an input triggers neuron output lying outside the output range
decided by the training data. That is, if an input is more dissimilar to
training data, it is more likely to increase NBC/SNAC. Also, DSC increases the
coverage if an input has a large distance (measured using DSA) to neuron outputs
of training data.
\subsection{Guiding Input Mutation in DNN Testing}
\label{subsec:rq2}

This section uses existing criteria and \tool\ as the objective to guide input
mutation. This denotes a typical feedback-driven (fuzz) testing setting, where
inputs are mutated to maximize the feedback yielded by the criteria. We
evaluate the mutated inputs on 1) \#triggered faults, 2) naturalness, and 3)
diversity of the erroneous behaviors.

\begin{figure}[!ht]
  \centering
  \vspace{-5pt}
  \includegraphics[width=0.70\linewidth]{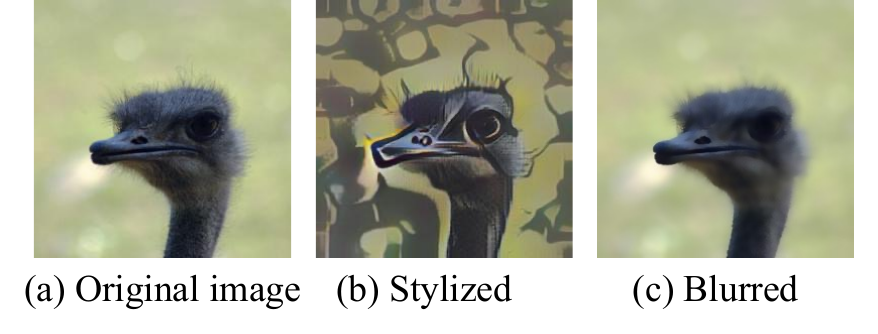}
  \vspace{-5pt}
  \caption{Stylizing and blurring.}
  \vspace{-15pt}
  \label{fig:style}
\end{figure}

\noindent \textbf{Setup.}~We use the same image classification models and
datasets as in \S~\ref{subsec:rq1}. \T~\ref{tab:op-rq2} and
\T~\ref{tab:config-rp2} list mutation methods and configurations of coverage criteria.
In sum, mutations in \T~\ref{tab:op-rq2} are representative, for
example, blurring and stylizing (see \F~\ref{fig:style}) reveal the
\textit{texture-bias} (i.e., rely on texture to make decisions) of
DNNs~\cite{geirhos2018imagenettrained,brendel2019approximating}.

We launch fuzzing using either prior coverage criteria or \tool\ as in
\A~\ref{alg:fuzz}. $terminate()$ returns \texttt{True} when reaching 10K
iterations or exceeding $6$ hours. $is\_valid()$ is adopted
from~\cite{xie2018coverage}, which deems a mutation as valid if 1) the
\#changed-pixels is less than $\alpha \times$ \#pixels or 2) the maximum of
changed pixel value is less than $\beta \times 255$. We set $\alpha$ to $0.2$
and $\beta$ to $0.4$. As a baseline, we do not check validity and coverage
(line 7 in \A~\ref{alg:fuzz}) for random mutation. That is, line 7 in
\A~\ref{alg:fuzz} always yields true for random mutation. We randomly select
1K inputs from the test dataset as the fuzzing seeds.

\begin{table}[t]
  \begin{minipage}{0.65\linewidth}
  \centering
  \caption{Transformations.}
  \vspace{-5pt}
  \label{tab:op-rq2}
  \resizebox{1.00\linewidth}{!}{
  \begin{tabular}{
    @{\hspace{1pt}}l@{\hspace{1pt}}|
    @{\hspace{1pt}}c@{\hspace{1pt}}|
    @{\hspace{1pt}}c@{\hspace{1pt}}
    }
    \hline
     Type         & Operator & Remark \\
    \hline
    Pixel-Level   & contrast, brightness & Robustness  \\
    \hline
    Affine        & translation, scaling, rotation & Shape bias \\
    \hline
    Texture-Level & blurring, stylizing   & Texture bias  \\
    \hline
  \end{tabular}
  }
\end{minipage}%
\begin{minipage}{0.35\linewidth}
  \centering
  \caption{Config.}
  \vspace{-5pt}
  \label{tab:config-rp2}
  \resizebox{1.0\linewidth}{!}{
  \begin{tabular}{
    @{\hspace{1pt}}c@{\hspace{1pt}}|
    @{\hspace{1pt}}c@{\hspace{1pt}}|
    @{\hspace{1pt}}c@{\hspace{1pt}}
  }
  \hline
  DSC: $0.1$ & \multicolumn{2}{c}{NC: $0$} \\
  \hline
  MDSC: $10$ & \multicolumn{2}{c}{LSC: $10$} \\
  \hline
  TKNC: $10$ & \multicolumn{2}{c}{KMNC: $100$} \\
  \hline
  TKNP: $50$ & \multicolumn{2}{c}{CC: $10/1000$} \\
  \hline
  \end{tabular}
  }
\end{minipage}
\vspace{-5pt}
\end{table}

\begin{algorithm}[t]
\footnotesize
\caption{Fuzzing algorithm.}
\label{alg:fuzz}
\SetKw{KwBy}{by}
  Transformation Set: $\mathcal{T}$; Fuzzing Seed: $S$\;
  Tested DNN: $D$; Criterion: $C$;
  $num \gets 50$\;
  \While{\textbf{not} terminate()}{
      $s = sample(\mathcal{S})$\;
      \For{$i\gets0$ \KwTo $num$ \KwBy $1$}{
          $t = sample(\mathcal{T})$;
        $\hat{s} \gets t(s)$\; 
          \If{$is\_valid(\hat{s}, s)$ \textbf{and} $coverage\_inc(C, D)$}{
              $S.add(\hat{s})$;
              $update\_coverage(C)$\;
              \textbf{break}\;
          }
      }
  }
\end{algorithm}

\begin{table}[t]
  \caption{Fuzzing results. Faults, Outputs, Entropy denote triggered faults, fuzzing outputs,
   scaled entropy, respectively.}
   \vspace{-5pt}
  \label{tab:fuzz-rq2}
  \centering
\resizebox{0.95\linewidth}{!}{
  \begin{tabular}{
    @{\hspace{1pt}}l@{\hspace{1pt}}|
    @{\hspace{1pt}}c@{\hspace{1pt}}|
    @{\hspace{1pt}}c@{\hspace{1pt}}|
    c|c|c|c
    }
    \hline
        \multicolumn{2}{c|}{} & \#Faults/\#Outputs & IS & FID & \#Classes & Entropy \\
    \hline
    \multirow{12}{*}{CIFAR10} & Rand & 4,730/\textbf{10,000} & 5.08 & 185.59 & 10 & 0.75  \\
    & NC & 1,063/1,817 & 4.65 & 176.57 & 10 & 0.94 \\
    & KMNC & 0/0 & N/A & N/A & 0 & 0  \\
    & NBC & 8/10 & 1.99 & 415.01 & 4 & 0.91  \\
    & SNAC & 5/8 & 2.70 & 469.29 & 1 & 0  \\
    & TKNC & 844/1,458 & 5.36 & 183.73 & 10 & 0.93  \\
    & TKNP & 5,983/9,380 & 5.30 & 170.19 & 10 & 0.85  \\
    & CC & 37/52 & 2.55 & 358.93 & 10 & 0.92  \\
    & LSC & 73/93 & 2.64 & 350.67 & 9 & 0.84  \\
    & DSC & 743/889 & 4.45 & 234.68 & 10 & 0.85  \\
    & MDSC & 210/284 & 2.96 & 279.24 & 9 & 0.58  \\
    & \textbf{\tool} & \textbf{8,716}/\textbf{10,000} & \textbf{5.61} & \textbf{167.87} & \textbf{10} & \textbf{0.97} \\
    \hline
    \multirow{12}{*}{ImageNet} & Rand & 6,448/\textbf{10,000} & 8.05 & 131.85 & 595 & 0.82 \\
    & NC &  3,309/4,157 & 8.05 & 99.35 & 672 & 0.91 \\
    & KMNC & 0/0 & N/A & N/A & 0 & 0  \\
    & NBC &  5/6 & 1.85 & 367.84 & 4 & 0.96   \\
    & SNAC &  8/10 & 2.62 & 372.49 & 7 & 0.97 \\
    & TKNC &  3,127/3,967 & 7.82 & 99.87 & 668 & 0.90 \\
    & TKNP &  7,190/\textbf{10,000} & 7.77 & 101.08 & 711 & 0.87 \\
    & CC &  7,902/\textbf{10,000} & 7.13 & 102.84 & 708 & 0.86 \\
    & LSC &  7,246/\textbf{10,000} & 7.74 & 100.38 & 700 & 0.79 \\
    & DSC &  24/31 & 2.54 & 278.55 & 17 & 0.93 \\
    & MDSC &  N/A & N/A & N/A & N/A & N/A \\
    & \textbf{\tool} &  \textbf{9,975}/\textbf{10,000} & \textbf{8.14} & \textbf{98.87} & \textbf{732} & 0.88* \\
    \hline
  \end{tabular}
  }
  \begin{tablenotes}
    \footnotesize
    \item *Higher Entropy is better if two \#Classes are equal, otherwise, higher \#Classes is better.
\end{tablenotes}
\vspace{-10pt}
\end{table}

\noindent \textbf{Triggered Faults.}~The \#triggered faults of ResNet trained on CIFAR10
and ImageNet are presented in \T~\ref{tab:fuzz-rq2}.
DSC is unsuitable to be an objective, because its computing cost impedes fuzzing
throughput (the fuzzer mutates images for only a few times within 6 hours). Images mutated under the guidance of \tool\
(last row) have the highest error triggering rates compared with random mutation
(3rd row) and other criteria, especially NBC, SNAC, KMNC. Recall that these criteria
denote a neuron as activated if its output is higher than a threshold or is
out-of-range. Since modern DNNs frequently perform normalization, neuron outputs tend
to concentrate to certain regions (as reflected in \F~\ref{fig:visual}). Without
fine-grained feedbacks such as gradients, mutated images can hardly flip an
unactivated neuron into activated. 
Nevertheless, we find that the degree of
entanglement among neurons can vary w.r.t.~mutated inputs. Thus, \tool\ is very
effective to serve the fuzzing guidance, achieving the highest \#faults and
fault-triggering rates.
TKNP and CC have decent performance, since the patterns and clusters in TKNP and CC,
to some extent, can reflect the neuron entanglement.

\begin{figure*}[!ht]
  \centering
  \vspace{-5pt}
  \includegraphics[width=0.85\linewidth]{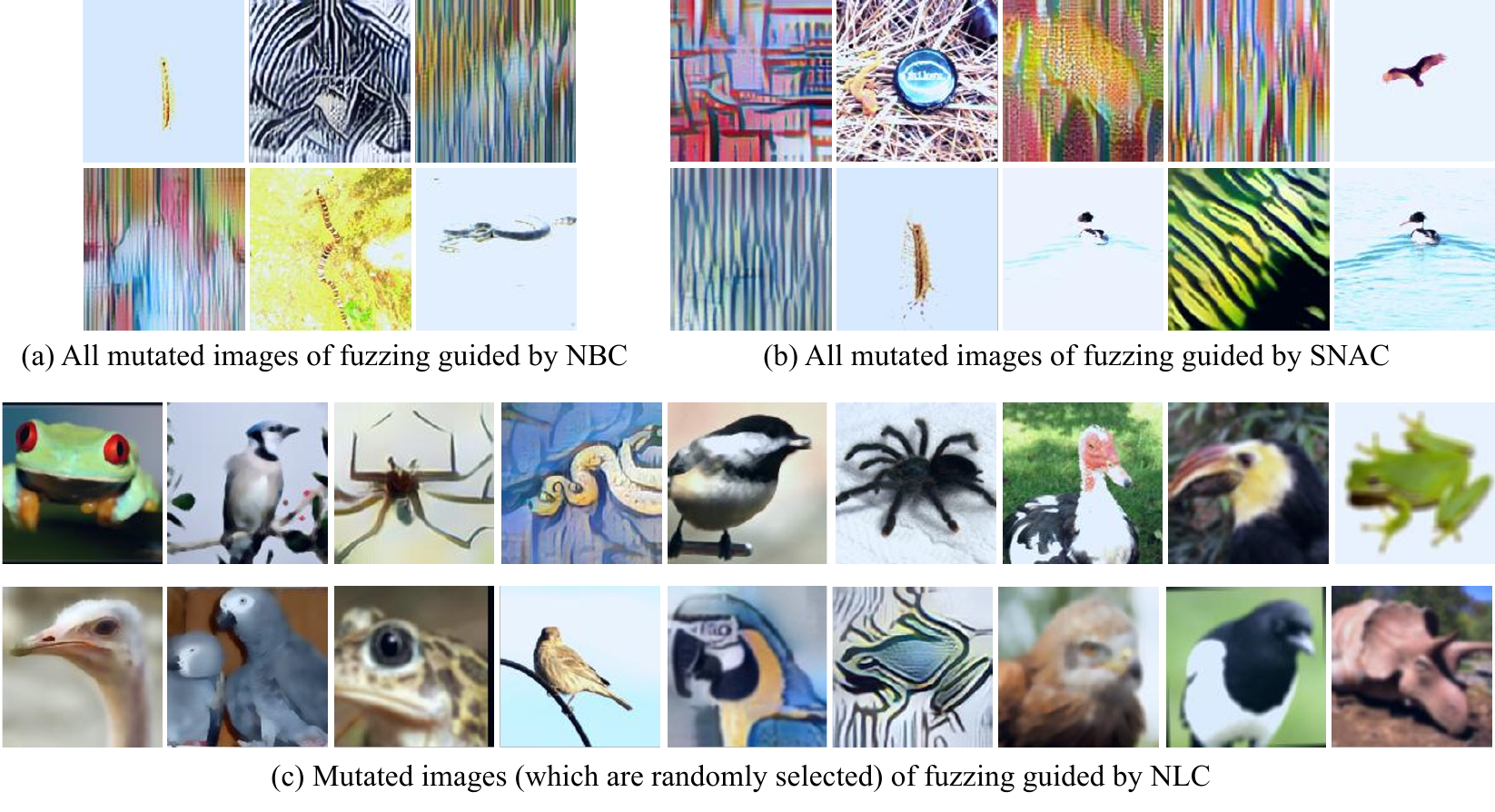}
  \vspace{-5pt}
  \caption{Mutated images of fuzzing guided by different criteria.}
  \vspace{-5pt}
  \label{fig:fuzz}
\end{figure*}

\noindent \textbf{Naturalness of Mutations.}~Contemporary testing works have pointed out the
importance of preserving naturalness of mutated images~\cite{harel2020neuron}.
We measure the naturalness of mutated inputs under each criterion; a good
criterion should lead to mutations that mostly preserve the naturalness.
We use Inception Score (IS)~\cite{salimans2016improved} and Fr$\grave{e}$chet
Inception Distance (FID)~\cite{heusel2017gans}, which are widely used in the AI
and SE community to assess the naturalness of images, as the metrics. A higher IS
score is better, whereas a lower FID score indicates better naturalness. The
evaluation results are also listed in \T~\ref{tab:fuzz-rq2}.

Random mutation's IS and FID scores are the baselines. Overall, mutated images
guided by \tool\ have the \textit{best scores} for both IS and FID. Mutated
images guided by SCs, NBC, SNAC have relatively low naturalness. SCs prefer
inputs that are distinct to training data (also pointed out in \S~\ref{subsubsec:diversity}).
Given the high comprehensiveness of
real-world datasets, SCs may likely push to produce unrecognizable images. NBC
and SNAC denote a neuron as activated if it has \textit{out-of-range} outputs.
Thus, its guided mutations tend to largely modify pixel values. We find that the
mutated images become less recognizable; \F~\ref{fig:fuzz} shows their mutated outputs
on ImageNet.
Overall, since neurons jointly process features in an image, mutations guided by
\tool, which aims to cover different neuron entanglements, are holistically
``bounded'' by features --- the mutated images (see \F~\ref{fig:fuzz})
thus have better visual quality.

\noindent \textbf{Diversity of Erroneous Behaviors.}~A higher diversity of
erroneous behaviors indicates uncovering a practically larger vulnerability
surface of DNN models. Overall, a collection of fault-triggering images are
regarded as more diverse if they cover more classes. In case that \#covered
classes is equal, we further use the scaled entropy, $-\frac{1}{|C|}\sum
p_{c}\log p_{c}$ where $p_{c}$ is the ratio of incorrect outputs predicted as
class $c$ and $|C|$ is the \#classes, to assess the diversity: a higher entropy
is better. The results are also presented in \T~\ref{tab:fuzz-rq2}. It is seen
that fault-triggering images generated by using \tool\ as the guidance cover the
most classes. We thus envision equipping fuzz testing frameworks with \nac\ to
likely identify more and diverse defects from real-world critical DNN systems.

\section{Discussion}
\label{sec:discussion}

\begin{table}[t]
    \caption{Time cost for 10K CIFAR10 inputs on ResNet.}
    \vspace{-5pt}
    \label{tab:time}
    \centering
  \resizebox{0.95\linewidth}{!}{
    \begin{tabular}{
                   c@{\hspace{2pt}}|
    @{\hspace{2pt}}c@{\hspace{2pt}}|
    @{\hspace{2pt}}c@{\hspace{2pt}}|
    @{\hspace{2pt}}c@{\hspace{2pt}}|
    @{\hspace{2pt}}c@{\hspace{2pt}}|
    @{\hspace{2pt}}c@{\hspace{2pt}}|
    @{\hspace{2pt}}c@{\hspace{2pt}}|
    @{\hspace{2pt}}c@{\hspace{2pt}}|
    @{\hspace{2pt}}c@{\hspace{2pt}}|
    @{\hspace{2pt}}c@{\hspace{2pt}}|
    @{\hspace{2pt}}c
    }
      \hline
      NC & KMNC & NBC & SNAC & TKNC & TKNP & CC & LSC & DSC & MDSC & \textbf{\tool} \\
      \hline
      2.3s & 8.3s & 2.4s & 2.3s & 3.6s & 431.5s & 75.9s & 37.3s & 432.5s & 14.3s & 4.4s \\
      \hline
    \end{tabular}
    }
   \vspace{-10pt}
\end{table}

\noindent \textbf{Time/Space Cost of \tool.}~\T~\ref{tab:time} compares the time
cost of \tool\ with previous criteria. \tool\ is slower than NC/NBC/SNAC/TKNC due
to covariance computation. It is slightly faster than KMNC/MDSC and much faster
than TKNP/CC/LSC/DSC. 

Despite both MDSC~\cite{kim2020reducing} and \tool\ use covariance matrix,
\tool\ requires less memory space. Let $m_i$ be \#neurons in the $i$-th DNN
layer and $K$ be \#classes in training data, \tool\ takes $\sum_i m^2_i$ space,
while MDSC takes $K\times(\sum_i m_i)^2$. To interpret the comparison: first,
$K$ is large in real-world datasets ($K=1000$ in ImageNet), hence MDSC may be
hardly used on ImageNet-trained DNNs but \tool\ can. Second, when $K$ is $1$, \tool\ 
still takes less space and saves more space for deeper DNNs.

\noindent \textbf{Interpretability and Usage.}~We view that DNN coverage
criteria (including \tool\ and prior works) may share common limitation of
shallow interpretability on their outcomes; For instance, as noted in
\S~\ref{subsec:nc}, NC simplifies the neuron output (a continuous value) to a
binary state (activated vs. unactivated). Hence, given a case of 100\% NC
coverage, we are concerned that different neuron output values and the
combination states of neurons/layers are far from fully examined. Worse yet,
once it hits 100\% (merely CIFAR10 training data can have 100\% coverage on
ResNet using $\langle \text{NC}, T=0.2 \rangle$), further tests cannot be guided
by NC. That is, ``saturating'' state of DNN criteria should not be interpreted
as revealing all DNN behaviors. Similarly, zero coverage reported by some
criteria does not imply that DNN behaviors are never explored. In
\T~\ref{tab:disc-rq1}, when using the same inputs (CIFAR10), NBC has zero
coverage, whereas $\langle \text{TKNP}, K=10 \rangle$ is saturated (increases
10,000 with 10,000 inputs).

High-level speaking, DNN coverage is \textit{not} simply analogous to software
coverage, where each statement is either ``covered'' or not. Outputs of DNN
coverage criteria depend on how those criteria, not merely the DNNs, are
designed/implemented. Fundamentally, neuron outputs are continuous, and neurons
in DNN layers collaborate to comprehend the semantics of high-dimensional inputs
(e.g., images). Explaining DNN decisions is still an open problem. Thus, DNN
coverage is not as clear as traditional software coverage, where we simply
``count'' if a statement/path/basic block is covered or not.

In sum, aligned with recent criteria like CC~\cite{odena2018tensorfuzz}, we do
not strive to establish a ``maximal value'' for \tool\ (since misprediction
cannot be eliminated currently). \textit{To use \tool, developers should try
their best to maximize \tool\ values with test inputs}. Also, it is infeasible
to universally decide the number of inputs required to maximize \tool, as this
number depends on the DNN structures, weights, and training data.

\section{Conclusion}
\label{sec:conclusion}

Given observations that DNNs approximate distributions, we propose eight design
requirements for DNN testing criteria. We design \tool, a criterion that
fulfills all requirements. \tool\ considers the continuity, correlation, distribution,
and density of layer outputs in a DNN. It also enables various optimizations.
We show that \tool\ can facilitate test input assessment and serve as guidance of
fuzz testing with better performance.

\bibliographystyle{plain}
\bibliography{bib/main}

\end{document}